\documentclass[10pt,twocolumn,letterpaper]{article}
\PassOptionsToPackage{table}{xcolor}
\usepackage[pagenumbers]{cvpr}

\usepackage{pifont}
\usepackage{wrapfig}
\usepackage{listings}
\usepackage{textcomp}
\lstdefinestyle{plannerprompt}{
  basicstyle=\fontencoding{T1}\ttfamily\footnotesize,
  columns=fullflexible,
  keepspaces=true,
  showstringspaces=false,
  breaklines=true,
  breakatwhitespace=true,
  breakindent=1em,
  upquote=true,
  frame=single,
  framerule=0.3pt,
  framesep=6pt,
  xleftmargin=7pt,
  xrightmargin=7pt,
  aboveskip=8pt,
  belowskip=12pt,
  captionpos=t
}

\definecolor{cvprblue}{rgb}{0.21,0.49,0.74}
\usepackage[breaklinks,colorlinks,allcolors=cvprblue,linktoc=all]{hyperref}
\title{Asynchronous Reasoning and Control for Efficient Robot Manipulation}

\author{Sombit Dey \qquad Luc Van Gool \qquad Danda Pani Paudel\\
INSAIT, Sofia University St. Kliment Ohridski\\
{\tt\small \{sombit.dey,luc.vangool,danda.paudel\}@insait.ai}}
\hypersetup{pdftitle={\thetitle},pdfauthor={Sombit Dey, Luc Van Gool, Danda Pani Paudel}}

\begin{document}
\addtocontents{toc}{\protect\setcounter{tocdepth}{-1}}
\twocolumn[{%
\renewcommand\twocolumn[1][]{#1}%
\maketitle
\begin{center}
\begin{minipage}[c]{0.56\textwidth}\centering
\includegraphics[width=\linewidth]{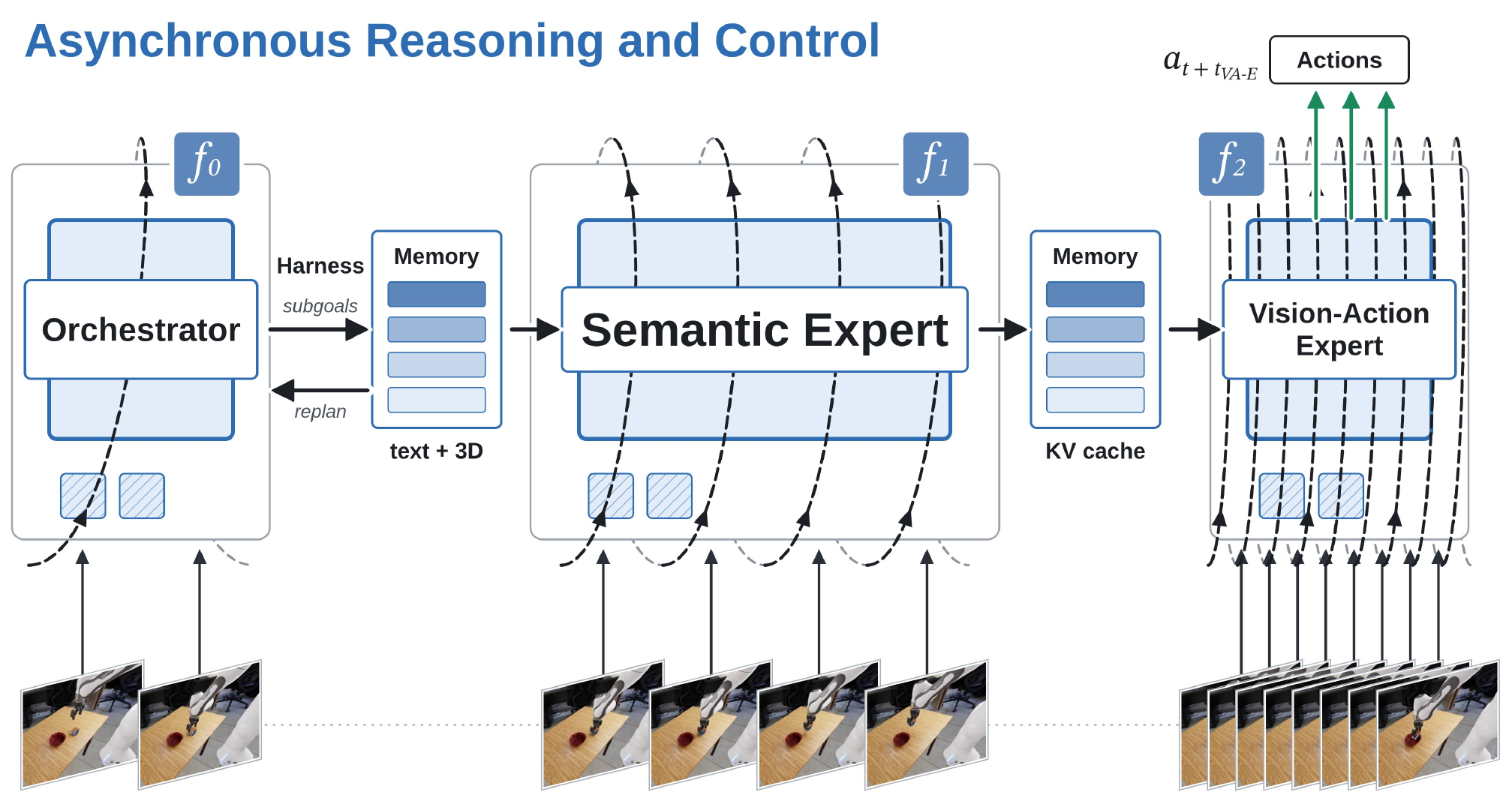}
\end{minipage}\hfill
\begin{minipage}[c]{0.43\textwidth}\centering
\includegraphics[width=\linewidth]{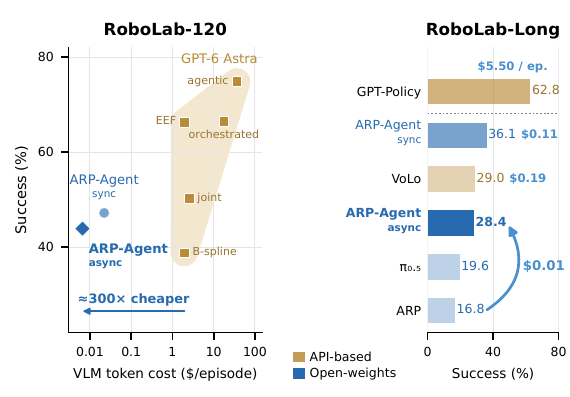}
\end{minipage}

\captionof{figure}{\textbf{Asynchronous reasoning and control.} \emph{Left:} ARP-Agent plans, updates semantic memory, and executes actions at separate rates. \emph{Right:} Task success and VLM token cost per episode on RoboLab-120 and RoboLab-Long. The $\approx300\times$ arrow compares GPT-6 Astra's reported direct end-effector policy cost on successful episodes with our local planner's API-equivalent cost across all episodes under a different simulator protocol (Table~\ref{tab:harness_efficiency}). On RoboLab-Long, asynchronous ARP-Agent raises success from 16.8\% to 28.4\% at an estimated planner cost of \$0.01 per episode (Table~\ref{tab:robolab_long_leaderboard_compact}).}
\label{fig:teaser}
\end{center}%
}]
\begin{abstract}
Robot manipulation requires both semantic reasoning and timely responses to a changing physical world. Large vision--language models provide the former, but their inference latency can limit the latter when control and task execution must wait for model updates. We present an asynchronous approach that allows reasoning and execution to proceed at independent rates, addressing the resulting challenge of acting on delayed information. At the policy level, Asynchronous Robot Policies (ARP) pair a semantic VLM with a lightweight vision--action expert. The expert combines fresh observations with the VLM's latest available KV-cache memory and is trained to account for memory staleness. At the task level, ARP-Agent extends this principle to long-horizon manipulation: skills continue executing while a planner reasons, and incoming plan revisions are reconciled with ongoing task progress. The system operates locally using publicly available model weights, without external model API calls. Experiments demonstrate improved ARP performance under unexpected object motion and on repeated real-world object transfers. Asynchronous ARP-Agent raises RoboLab-120 success from 25.2\% with standalone ARP to 43.9\% and achieves 28.4\% success on RoboLab-Long. Across RoboLab-120 and RoboLab-Long, asynchronous orchestration reduces planning expenditure while yielding lower success than blocking execution, exposing a trade-off between planning resources and task completion.%
\iftoggle{cvprfinal}{\space The project website is at \href{https://arp.insait.ai}{arp.insait.ai}.}{}

\end{abstract}
\section{Introduction}

Vision--language models (VLMs) bring semantic understanding and language grounding to robot manipulation, supporting both learned action policies and task-level planning~\citep{zitkovich2023rt2, kim2024openvla, saycan2022arxiv}. Physical interaction, however, requires timely responses to changing observations. An object may move during a reach, a grasp may slip, or contact may alter the scene while a model is computing. When action prediction depends on a new VLM inference, the cost of semantic processing constrains how quickly the policy can incorporate fresh observations. At the task level, a blocking planner similarly delays execution while it reasons about the next operation.

Semantic reasoning, task planning, and motor control need not be updated together. A semantic representation can remain useful across several action predictions, and an executable plan can support progress while a revision is computed. This suggests organizing robot decision-making as interacting processes that run at independent rates and communicate through persistent, timestamped state. The central challenge is then temporal coordination: how should a controller use semantic information from an earlier observation, and how should an executor apply a plan after the scene and task progress have changed?

We address this challenge at two levels. Within a learned policy, \textbf{Asynchronous Robot Policies (ARP)} separate a semantic VLM from a lightweight vision--action expert. The VLM periodically updates a projected memory derived from its key--value (KV) cache. The action expert predicts chunks from current images and proprioception together with the latest available memory, allowing visual feedback to influence actions without waiting for each semantic update. We train the expert with temporally offset observation--memory pairs and condition it on memory staleness. This training prepares it to combine current sensory evidence with delayed semantic context.

Across planning and skill execution, \textbf{ARP-Agent} coordinates a VLM planner, visual perception, ARP, and geometric manipulation primitives. After initialization, the executor continues available skills while the planner generates revisions. Here, persistent task context consists of autoregressively generated text plans, perceived 3D scene state, and execution feedback. Incoming revisions are reconciled with completed operations and the active execution state before replacing pending work. ARP and ARP-Agent share a design principle: each process maintains context that execution can use between updates. Semantic memory guides control as observations change, while executable plans support task progress as new decisions are computed.

We evaluate on real-world Franka manipulation, dynamic-object grasping, RoboLab-120, and RoboLab-Long. ARP improves progress on repeated object-transfer tasks and success under unexpected target motion. Asynchronous ARP-Agent raises RoboLab-120 success from 25.2\% to 43.9\% and improves long-horizon completion over standalone ARP. Compared with blocking orchestration, it uses approximately $3.7\times$ less estimated planner computation on RoboLab-120, with a 3.3-percentage-point decrease in success. These results characterize a trade-off between planning resources and task completion, using locally hosted models without external model API calls.

Our contributions are:
\begin{itemize}
    \item \textbf{Asynchronous visuomotor control:} an architecture that combines a semantic VLM's KV-cache memory with fresh visual observations, trained to accommodate memory staleness.
    \item \textbf{Asynchronous task orchestration:} a manipulation system that continues skill execution during planning and reconciles delayed revisions with intervening task progress.
    \item \textbf{Evaluation across reaction, planning, and cost:} experiments with dynamic objects, real robots, and long-horizon benchmarks that examine task performance, delayed updates, and computational trade-offs using publicly available model weights and public robot training data.
\end{itemize}

\section{Related Work}
\label{sec:related_new}

\subsection{Robot Policies}

Vision--language--action models (VLAs) transfer pretrained visual and linguistic representations to robot control~\citep{zitkovich2023rt2, kim2024openvla}. Large robot datasets, including Open X-Embodiment, DROID, and BridgeData V2, support learning across tasks and embodiments~\citep{open_x_embodiment_rt_x_2023, khazatsky2024droid, walke2023bridgedata}. Within this framework, actions can be represented as discrete tokens~\citep{kim2024openvla, pertsch2025fast} or generated through diffusion and flow matching~\citep{octo_2023, black2024pi_0}. Dedicated action experts connect VLM representations to continuous motor control~\citep{li2024cogact, reuss2025flower, pi05, nvidia2025gr00tn1openfoundation}, while intermediate reasoning and spatial grounding provide additional context for action prediction~\citep{zawalski2024ecot, zhao2025cotvla, qu2025spatialvla}.

A complementary direction uses world-action models to connect action generation with video prediction or future latent states~\citep{kim2026cosmos, ye2026worldactionmodelszeroshot, bfl2026flux3action, chen2026lawam, zhou2024dino, assran2025v}. Our work retains a VLM semantic backbone, training from PaliGemma-3B~\citep{beyer2024paligemma} on DROID with LAP's language--action objective~\citep{zha2026lap}. We examine how its semantic context can support control between VLM updates.

\subsection{Asynchronous Inference and Control}

Maintaining responsive control requires accounting for the time between observing the scene and executing an action. Real-time chunking overlaps prediction with execution by conditioning on committed actions~\citep{black2025real, black2025training}; VLASH and delay-aware diffusion policies likewise address observation--execution misalignment~\citep{tang2025vlash, liao2025delay}. DynamicVLA combines a compact VLM and an action expert with continuous inference and action streaming for moving-object manipulation~\citep{xie2026dynamicvla}. Its action expert receives visual context through the VLM backbone. These approaches motivate efficient action-chunk execution, including the RTC variant evaluated with ARP.

A related architectural direction allows semantic processing and control to update independently. AsyncVLA and TIC-VLA explore this organization for navigation~\citep{hirose2026asyncvla, huang2026tic}, while AR-VLA uses refreshable vision--language prefixes with an autoregressive action expert~\citep{hu2026ar}. Training for delayed context appears in Acting While Understanding~\citep{yan2026acting} and cache-conditioned driving policies~\citep{li2026think}. DuoCore-FS connects slow and fast pathways through a latent buffer~\citep{zou2025duocore}. Separately, flow-matching AsyncVLA schedules refinement within action generation~\citep{jiang2026asyncflow}.

DAM-VLA extends asynchronous processing to visual, proprioceptive, and force/torque streams through modality-specific latent buffers~\citep{vanjani2026dam}. It evaluates fine-tuned X-VLA models on real-world tasks spanning visual guidance and contact-sensitive manipulation. ARP focuses on a fresh visual feedback pathway: its vision--action expert encodes current images independently of the semantic VLM and combines them with age-conditioned KV-cache memory. We study this policy-level organization alongside asynchronous planning and skill execution.

\subsection{Planning and Robot Orchestration}

At the task level, foundation models connect language instructions to executable behavior through skill selection~\citep{saycan2022arxiv}, multimodal reasoning~\citep{driess2023palme}, and programs that invoke perception and control tools~\citep{codeaspolicies2022, fu2025capx}. Modular systems combine open-vocabulary perception, geometric planning, and learned policies to carry out these decisions~\citep{shen2026tiptop, shi2025maestro, huang2026roboharness, chen2026show, 2026amp}. VoLo supports planning and recovery through interruptible VLA/WAM tools~\citep{chen2026volo}, while Astranomy studies frontier models that directly generate robot actions~\citep{yang2026astranomy}. ARP-Agent follows the modular approach, allowing skills to continue during planning and reconciling incoming revisions with intervening execution progress. Our evaluation examines how delayed updates affect task completion and planning cost.

\section{Method}
\label{sec:method}

Our method has two components: Asynchronous Robot Policy (ARP) for low-level control and ARP-Agent for task-level planning and execution. ARP separates semantic inference from visuomotor control; ARP-Agent applies asynchronous execution to manipulation skills.

\subsection{Asynchronous Robot Policy}
\label{sec:arp}

ARP has two independently scheduled components. A semantic expert, a VLM, periodically processes the instruction and a visual observation and exports its key--value (KV) cache as a semantic memory. A vision--action (VA) expert combines this memory with the current images and proprioception to predict chunks whose actions are executed at control rate $f$. The VA expert keeps acting on the latest available memory while the semantic expert computes an update.

Let $\tau(t)$ denote the observation step whose KV cache produced the latest memory available at control step $t$. The memory age, in control steps, is
\begin{equation}
    d_t=t-\tau(t).
    \label{eq:memory_staleness}
\end{equation}
The VA expert predicts an $H$-step action chunk
\begin{equation}
    a_{t:t+H-1}=\pi_{\mathrm{VA}}(I_t,s_t,M_{\tau(t)},d_t),
    \label{eq:policy_formulation}
\end{equation}
where $I_t$ are the current camera images, $s_t$ is the proprioceptive state, and $M_{\tau(t)}$ is the semantic memory. Each chunk prediction thus uses current observations, even when semantic memory is stale; the RoboLab executor uses $H=16$ and executes eight actions per chunk.

\begin{figure}[t]
    \centering
    \includegraphics[width=\linewidth]{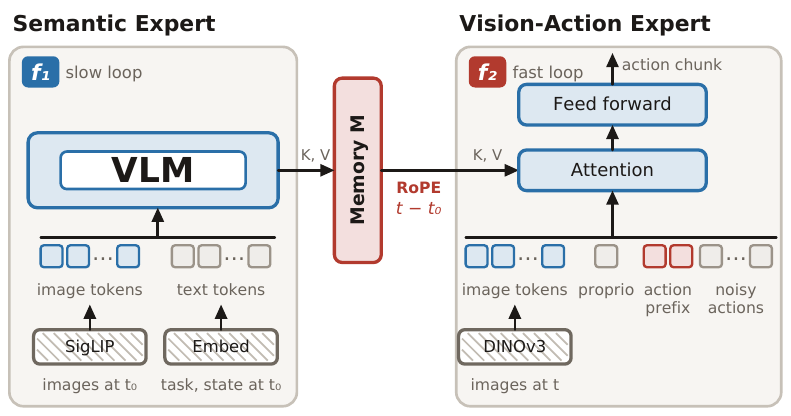}
    \caption{ARP reuses a VLM KV-cache memory while the vision--action expert acts on current observations.}
    \label{fig:arp_architecture}
\end{figure}

\paragraph{Semantic memory.}
The semantic expert is a PaliGemma-3B VLM~\citep{beyer2024paligemma} trained with LAP's language--action objective~\citep{zha2026lap}. At an update step $\tau$, it encodes the instruction and the camera images, and we tap the keys and values $KV_\tau$ of $L=8$ evenly spaced layers of its language model. A learned adapter maps them to a memory of fixed width,
\begin{equation}
    M_\tau=\operatorname{LNLinear}(KV_\tau)+P,
    \label{eq:cache_adapter}
\end{equation}
where $\operatorname{LNLinear}$ is layer normalization followed by a linear projection to the width of the VA expert and $P$ is a learned positional embedding; $M_\tau$ is then re-projected into one key--value pair for each layer of the VA expert. The memory does not encode its age, so it is computed once per update and held fixed. The age enters only when the memory is read: its keys are rotated by a temporal rotary embedding at time $-d_t$, while the VA expert's own tokens sit at time $0$, so attention can account for how old the memory is. This fixes the memory width presented to the VA expert; the number of memory tokens still follows the VLM prefix length (Fig.~\ref{fig:arp_architecture}).

\paragraph{Vision--action expert.}

The VA expert is a smaller transformer (Gemma-300M) whose tokens are visual tokens of the current images from its own DINOv3 encoder, a proprioception token from $s_t$, and $H$ action tokens; in every layer, these tokens attend jointly to the memory and to each other. Actions are generated by flow matching~\citep{lipman2024flow}. During training, the action tokens carry a noisy chunk $x_r=(1-r)\,a_{t:t+H-1}+r\,\epsilon$ with Gaussian noise $\epsilon$ and flow time $r\in(0,1]$, and the expert predicts the velocity $\epsilon-a_{t:t+H-1}$; at inference, it integrates from noise ($r=1$) to an action chunk ($r=0$). The flow time is embedded sinusoidally and modulates every layer through adaptive RMS normalization (AdaRMS); the memory age is not part of this conditioning. In the real-time chunking variant of Sec.~\ref{sec:analysis_arp}, the first action tokens instead carry the committed actions that execute while the new chunk is computed: they enter clean rather than noised, are marked by a learned prefix embedding, and are excluded from the loss.

\paragraph{Training with stale memory.}
Both experts are trained on DROID~\citep{khazatsky2024droid}, pairing each semantic memory with current and later observations from the same demonstration. Following the principle of knowledge insulation~\citep{driess2025insulating}, the action loss stops at the semantic memory boundary, while the semantic expert retains its language--action objective. Sampling and deployment schedules are given in Appendices~\ref{sec:appendix_arp} and~\ref{supp:inference_timing}, respectively.

\subsection{ARP-Agent: Asynchronous Policy Orchestration}
\label{sec:arp_agent}

ARP-Agent extends asynchronous execution to task orchestration. A locally hosted VLM planner (Qwen3-VL-32B~\citep{qwen3vl2025}) coordinates open-vocabulary perception (SAM~3~\citep{carion2026sam} with depth), ARP, and geometric manipulation primitives. Unlike ARP’s KV-cache memory, ARP-Agent maintains task context as autoregressive text plans and 3D scene state derived from perception. The executor continues the current plan while the VLM prepares a revision, then reconciles that revision with intervening task progress.

\paragraph{Perception and grounded execution.}
The harness maintains an object-centric scene representation from camera observations, calibrated depth, and robot proprioception. Planner-selected text queries identify objects and destinations; perception provides their geometry in the robot base frame. Perceptual predicates estimate whether an object is held and whether it has been placed in a destination, providing execution feedback without simulator success signals.

The planner specifies object-centric operations such as picking an object or placing it in, on, or beside a destination. The executor grounds these operations using perceived geometry and combines two complementary controllers: IK-based primitives perform geometric positioning and transport, while ARP handles grasping and release. Geometric primitives also provide fallbacks for cluttered scenes and unsuccessful policy execution. ARP receives current observations and the original task instruction; the executor guides it toward the intended operation by positioning the gripper near the selected object or destination. Execution outcomes update the task state and inform subsequent planning.

\paragraph{Stateless plan revision.}
Each planner call receives a self-contained snapshot comprising the instruction, annotated images, object geometry, running and pending operations, and recent execution outcomes. It returns perception queries, an ordered plan for the remaining task, and optional interruption and completion flags. Returning multiple operations per call allows execution to proceed between planning updates. The harness maintains persistent task state, so the planner does not require an accumulating conversation history.
\begin{figure}[t]
    \centering
    \includegraphics[width=\linewidth]{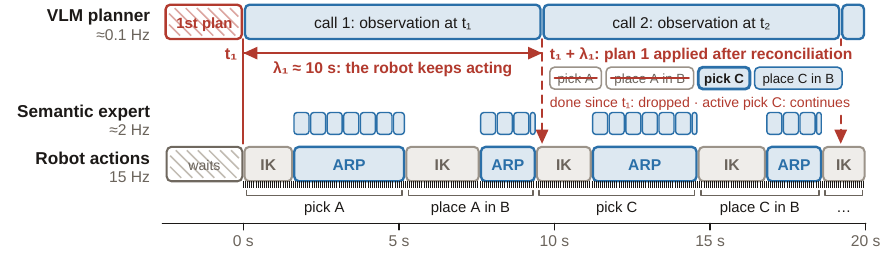}
    \caption{\textbf{ARP-Agent timeline (illustrative).} The robot keeps acting while the planner ($\approx$0.1\,Hz) revises the plan. A plan observed at $t_1$ is applied at $t_1{+}\lambda_1$ (Eq.~\ref{eq:plan_apply}), after operations completed since $t_1$ are dropped; the active operation continues. The semantic expert refreshes at $\approx$2\,Hz while ARP controls the arm, and actions run at 15\,Hz.}
    \label{fig:arp_agent_timeline}
\end{figure}
\paragraph{Reconciling delayed plans.}
After initialization, the executor follows the current plan while the planner computes a revision. A revision based on the observation at control step $t_k$ becomes available at
\begin{equation}
    t_k^{\mathrm{apply}}
    = t_k + \left\lceil f\lambda_k \right\rceil,
    \label{eq:plan_apply}
\end{equation}
where $\lambda_k$ is the measured planning latency and $f$ the control rate (Fig.~\ref{fig:arp_agent_timeline}). Before applying it, the harness drops operations completed since $t_k$ or incompatible with the current gripper state, to reduce repeated work and incompatible operations. The reconciled plan replaces the pending operations while the active one continues; an admissible interruption cancels only commands not yet dispatched. The next call uses the newest observation. When no operation is executable, the robot holds its position, and planner-declared completion ends execution only when the gripper is empty.

\paragraph{Asynchrony across policy and orchestration.}
Both levels separate the production of context from its consumption during execution. Within ARP, the action expert uses fresh observations and accounts for the age of its semantic memory. Within ARP-Agent, the executor continues available skills and reconciles incoming plans with intervening task progress. Together, these mechanisms allow semantic reasoning and task planning to proceed independently of ongoing robot control.

\section{Experiments and Results}

We evaluate the feasibility, task performance, and computational efficiency of asynchronous robot policies. We first examine manipulation with static and dynamic objects to assess task success and responsiveness to scene changes (Secs.~\ref{sec:franka} and~\ref{sec:dynamic_tasks}).
We further evaluate asynchronous orchestration of ARP-Agent on long-horizon tasks requiring multiple subgoals on the RoboLab-Long benchmark~\cite{ladder2026} (Sec.~\ref{sec:robolab_long}).
We also examine failure modes, semantic-memory staleness and age conditioning (Sec.~\ref{sec:ablations}), RTC action-chunk execution (Sec.~\ref{sec:analysis_arp}), and planner computation under delayed updates (Sec.~\ref{sec:orchestration_efficiency}).

\paragraph{Experimental Details}
We use the Franka robot as our primary embodiment and train on DROID~\citep{khazatsky2024droid}, a diverse collection of real-world manipulation trajectories, to evaluate ARP beyond task-specific training. After trajectory filtering, we train ARP with PaliGemma-3B~\citep{beyer2024paligemma} as the semantic backbone and DINOv3~\citep{simeoni2025dinov3} as the visual encoder for the vision--action expert. Following LAP~\citep{zha2026lap}, we jointly train the semantic backbone to predict language-based actions and the vision--action expert to generate action chunks using a flow-matching objective.
For ARP-Agent, we use Qwen3-VL-32B~\cite{qwen3vl2025} as the orchestrator. The orchestrator proposes object-centric operations, grounded by SAM3~\cite{carion2026sam} and depth perception; ARP receives the original task instruction. All learned components use open weights, and the harness operates without external model APIs or privileged simulator state.
We evaluate ARP and ARP-Agent in multiple settings: (a) RoboLab~\citep{yang2026robolab}, a photorealistic simulation benchmark enabling comparisons with generalist VLA baselines;
(b) Real-world Franka manipulation tasks on the DROID~\cite{khazatsky2024droid} hardware setup; and
(c) RoboLab-Long~\cite{ladder2026} to evaluate the capability of policies to handle multiple subgoals, follow language instructions, and divide tasks into subtasks.
Training, agent implementation, and evaluation details are provided in Appendices~\ref{sec:appendix_arp}, \ref{supp:arp_agent}, and~\ref{sec:franka_appendix}, respectively.

\subsection{Franka and RoboLab-120 Results}
\label{sec:franka}
\begin{figure}
    \centering
    \vspace{-15pt}
    \includegraphics[width=\linewidth]{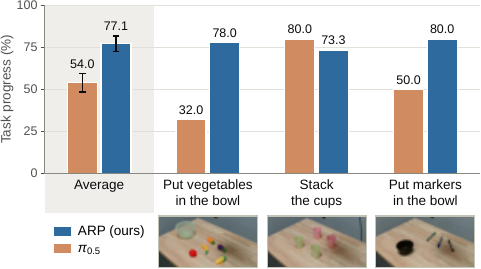}
    \caption{Real-world Franka evaluation: mean task progress over 10 rollouts per model per task.}

    \label{fig:franka_results}
    \vspace{-10pt}
\end{figure}
Figure~\ref{fig:franka_results} compares real-world Franka rollouts over three tasks, with 10 trials per model per task. ARP completes more grasp-and-place operations on the two object-transfer tasks within the fixed 90\,s budget, consistent with more robust repeated manipulation. On vegetables, it achieves approximately $2.4\times$ the task progress per allotted minute compared with $\pi_{0.5}$. The advantage also holds for markers, while $\pi_{0.5}$ scores higher on cup stacking.

On RoboLab-120, ARP achieves 25.2\% overall success, whereas the published $\pi_{0.5}$ leaderboard score is 28.0\% (Table~\ref{tab:robolab_leaderboard_compact}); simulator differences mean these scores provide context rather than a controlled comparison. ARP uses DROID alone for robot training, while $\pi_{0.5}$ benefits from broader and more diverse robot pretraining, including proprietary data~\cite{pi05}. ARP performs best on simple tasks and struggles on complex tasks. We hypothesize that the limited scale and task coverage of its DROID-only robot training contribute to this difficulty, motivating evaluation with broader training data.

Combining ARP with planning, perception, and geometric skills raises success from 25.2\% to 43.9\% in asynchronous ARP-Agent, with gains across all difficulty splits; blocking orchestration reaches 47.2\%. On complex tasks, success rises from 12.9\% to 31.2\%. This indicates that coordinating the policy with explicit planning and complementary manipulation skills helps address tasks that ARP alone struggles to complete. All components run locally using publicly available model weights and public robot training data, without external model API calls.

\providecommand{\cmark}{\textcolor{green!60!black}{\ding{51}}}
\providecommand{\xmark}{\textcolor{red}{\ding{55}}}
\begin{table*}[t]\centering\small
\caption{\textbf{RoboLab-120 leaderboard comparison.} Overall success rate (SR) and success by task difficulty, in percent, under default instructions. Model/Data indicate public availability of model weights and all robot training data, respectively: \cmark{} denotes available; \xmark{} denotes unavailable or undisclosed. External scores are from the November 15, 2026 leaderboard~\cite{robolabLeaderboard2026}}

\label{tab:robolab_leaderboard_compact}
\setlength{\tabcolsep}{6pt}
\begin{tabular}{llcrrrr}\toprule
 & & Public & & \multicolumn{3}{c}{SR by difficulty} \\
\cmidrule(lr){5-7}
Policy & Type & Model / Data & SR & Simple & Moderate & Complex \\ \midrule
AMP1~\cite{2026amp} & Agent & \cmark{} / \xmark & 86.0 & 86.8 & 87.7 & 79.1 \\
SenseNova-RoboRSI & Agent & \xmark{} / \xmark & 57.1 & 59.4 & 56.2 & 50.6 \\
Lakeside & LWM & \xmark{} / \xmark & 51.0 & 57.6 & 44.1 & 42.4 \\
BiMind v0.5 & VLA & \xmark{} / \xmark & 49.2 & 51.6 & 52.6 & 32.4 \\
Praxis-1 & WAM & \cmark{} / \xmark & 46.2 & 46.9 & 47.9 & 40.0 \\
FLUX 3 Action~\cite{bfl2026flux3action} & WAM & \cmark{} / \xmark & 42.9 & 49.1 & 39.2 & 28.2 \\
Cosmos3-Nano-Policy~\cite{cosmos3} & WAM & \cmark{} / \xmark & 36.8 & 40.6 & 35.4 & 25.3 \\
VoLo~\cite{chen2026volo} & Agent & \cmark{} / \xmark & 28.2 & 32.7 & 25.6 & 17.6 \\
$\pi_{0.5}$~\cite{pi05} & VLA & \cmark{} / \xmark & 28.0 & 29.7 & 31.5 & 13.5 \\
DreamZero~\cite{ye2026worldactionmodelszeroshot} & WAM & \cmark{} / \cmark & 25.7 & 26.1 & 30.0 & 14.1 \\
Cosmos3-Edge-Policy~\cite{cosmos3} & WAM & \cmark{} / \xmark & 22.9 & 25.6 & 23.3 & 11.8 \\
$\pi_0$-FAST~\cite{pertsch2025fast} & VLA & \cmark{} / \xmark & 15.5 & 20.2 & 13.3 & 2.9 \\
GR00T N1.6~\cite{nvidia2025gr00tn16} & VLA & \cmark{} / \xmark & 7.2 & 8.8 & 7.9 & 0.0 \\
$\pi_0$~\cite{black2024pi_0} & VLA & \cmark{} / \cmark & 5.0 & 7.2 & 3.6 & 0.0 \\
\midrule
ARP & Policy & \cmark{} / \cmark & 25.2 & 31.4 & 20.5 & 12.9 \\
ARP-Agent (Sync) & Agent & \cmark{} / \cmark & 47.2 & 57.3 & 39.0 & 27.6 \\
ARP-Agent (Async) & Agent & \cmark{} / \cmark & 43.9 & 49.8 & 39.7 & 31.2 \\
\bottomrule\end{tabular}
\end{table*}

\subsection{Dynamic Tasks}
\label{sec:dynamic_tasks}

\begin{wraptable}{r}{0.48\columnwidth}
\centering
\small
\setlength{\tabcolsep}{3pt}
\caption{\textbf{Dynamic grasping.} Success (\%) at perturbation probabilities $p$. Higher is better.}
\label{tab:dyn_bump}
\begin{tabular}{lcc}
\toprule
\textbf{Policy} & $p=0.5$ & $p=1$ \\
\midrule
$\pi_0$ & 20.0 & 12.7 \\
ARP & \textbf{46.7} & \textbf{44.7} \\
\bottomrule
\end{tabular}
\end{wraptable}

We evaluate ARP-1B and a $\pi_0$-style 1B baseline on grasping moving objects among static distractors; these use a separate simulation configuration from the DROID-trained policy. A velocity perturbation, scheduled with probability $p$, unexpectedly changes the target's motion during the reach. Both policies are trained on the same synthetic demonstrations and evaluated at $p=0.5$ and $p=1$. Dataset, perturbation, and evaluation details are provided in Appendix~\ref{supp:dyn_bump_details}.

As shown in Table~\ref{tab:dyn_bump}, ARP achieves 44.7\% success at $p=1$, compared with 12.7\% for $\pi_0$, a gain of 32.0 percentage points. ARP also performs better at $p=0.5$ and shows a smaller decline as perturbations become more frequent. These results support improved robustness to unexpected target motion. A qualitative comparison is provided in Appendix~\ref{supp:dyn_bump_details}.

\subsection{Long-Horizon Manipulation Tasks}
\label{sec:robolab_long}
\providecommand{\cmark}{\textcolor{green!60!black}{\ding{51}}}
\providecommand{\xmark}{\textcolor{red}{\ding{55}}}
\providecommand{\accOpen}{\textcolor{green!50!black}{\textbf{Open}}}
\providecommand{\accAPI}{\textcolor{red!80!black}{\textbf{API}}}
\begin{table*}[t]\centering\small
\caption{\textbf{RoboLab-Long results}~\cite{ladderLeaderboard2026} on 44 tasks with eight episodes per task. Success rate (SR) and subgoal completion are percentages; task length is the number of subgoals. Open/API indicate local public weights or closed-model API access. ARP-Agent costs are API-equivalent planner estimates.}
\label{tab:robolab_long_leaderboard_compact}
\setlength{\tabcolsep}{6pt}
\begin{tabular}{llcrrrrrrr}\toprule
 & & & & & \multicolumn{4}{c}{SR by task length} & \\
\cmidrule(lr){6-9}
Policy & Type & Access & SR\% & Subgoals\% & 1--2 & 3--4 & 5--7 & 8--10 & VLM cost/ep. (\$) \\ \midrule
GPT-Policy & Agent & \accAPI & 62.8 & 78.7 & 93.8 & 78.4 & 51.0 & 31.8 & 5.501 \\
PSP-Agent & Agent & \accAPI & 51.4 & 81.6 & 81.2 & 64.8 & 42.7 & 20.5 & 0.196 \\
VoLo & Orchestrator & \accAPI & 29.0 & 62.0 & 73.8 & 34.1 & 9.4 & 4.5 & 0.191 \\
CaP-X & Agent & \accAPI & 20.7 & 42.4 & 55.0 & 25.0 & 3.1 & 4.5 & 0.061 \\
$\pi_{0.5}$ & Policy & \accOpen & 19.6 & 56.6 & 57.5 & 18.2 & 5.2 & 2.3 & 0.000 \\
TiPToP & Agent & \accAPI & 17.3 & 22.4 & 48.8 & 22.7 & 2.1 & 0.0 & 0.006 \\
\midrule
ARP & Policy & \accOpen & 16.8 & 49.8 & 57.5 & 13.6 & 1.0 & 0.0 & 0.000 \\
ARP-Agent (Sync) & Agent & \accOpen & 36.1 & 64.3 & 83.8 & 44.3 & 15.6 & 6.8 & 0.113 \\
ARP-Agent (Async) & Agent & \accOpen & 28.4 & 62.1 & 80.0 & 34.1 & 6.3 & 0.0 & 0.010 \\
\bottomrule\end{tabular}
\end{table*}

We evaluate ARP and ARP-Agent on RoboLab-Long~\cite{ladder2026}, with 44 tasks and eight episodes per task (Table~\ref{tab:robolab_long_leaderboard_compact}). Asynchronous ARP-Agent increases full-task success from 16.8\% for standalone ARP to 28.4\%, and subgoal completion from 49.8\% to 62.1\%. These gains measure the combined contribution of planning, perception, and hybrid skill execution.

Blocking orchestration achieves 36.1\% success and 64.3\% subgoal completion. Asynchronous execution reduces the estimated planner cost from \$0.113 to \$0.010 per episode, but loses 7.7 percentage points of success and completes no eight-to-ten-subgoal episodes. The cost estimates cover planner tokens, excluding perception and action-policy computation. Protocol differences and accounting details are provided in Appendix~\ref{supp:orchestration_efficiency}.

Figure~\ref{fig:long_horizon_example} illustrates how planning, perception, geometric primitives, and ARP cooperate in a five-subgoal task. The planner revises object descriptions after failed localization, geometric primitives handle a cluttered grasp, and ARP recovers from a missed grasp and an object slip. This successful rollout illustrates the roles of the components; it does not isolate the effect of asynchronous scheduling.

\begin{figure*}[t]
    \centering
    \includegraphics[width=\textwidth]{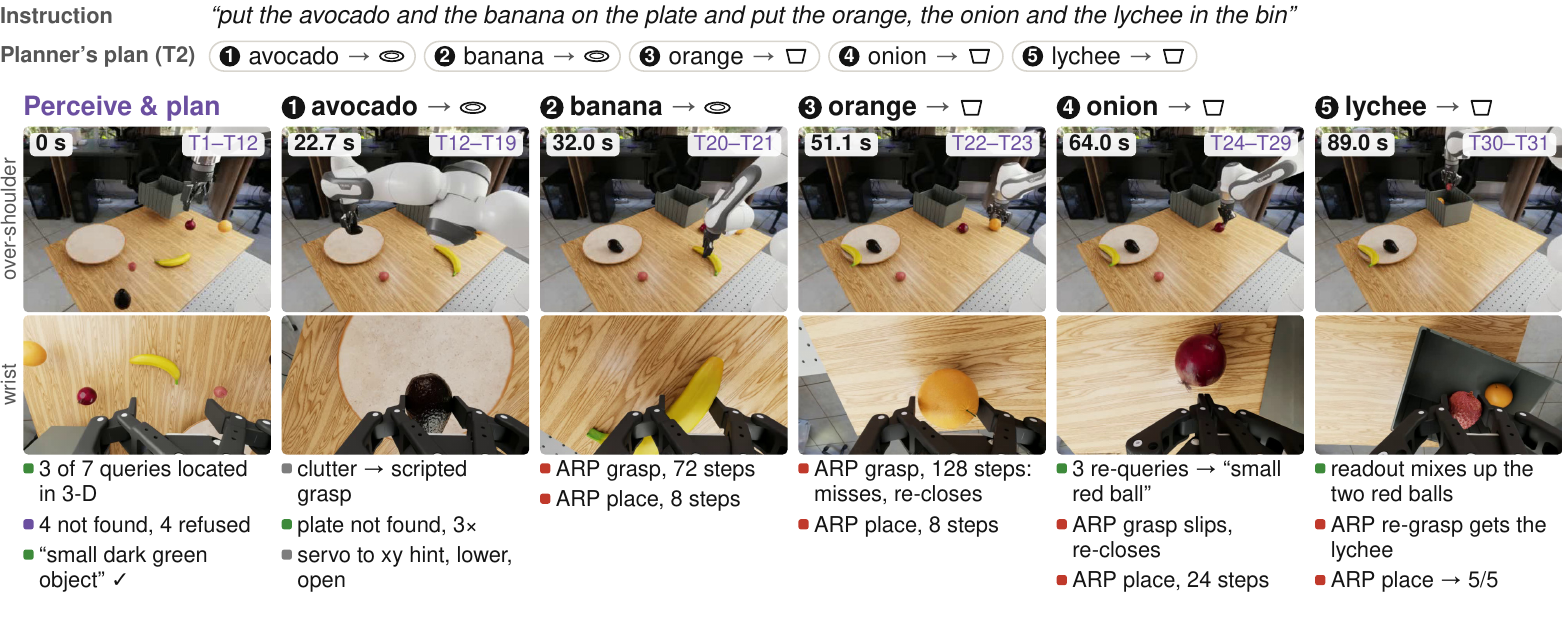}
    \caption{\textbf{Five-subgoal manipulation rollout.} Columns show perception and planning followed by manipulation of five objects, read from left to right; rows show the over-shoulder and wrist views. Plate and bin icons indicate destinations. T denotes planner turn, and timestamps denote simulated time at the selected snapshots. Colored squares identify planning (purple), perception (green), geometric control (gray), and ARP (red). The episode completes in 89.4\,s of simulated time; the final snapshot is at 89.0\,s. ARP controls 437 of 1,341 steps, including recovery from a missed grasp and an object slip. Simulated time does not establish wall-clock execution speed.}
    \label{fig:long_horizon_example}
\end{figure*}

\begin{table}[!t]
\centering
\small
\setlength{\tabcolsep}{3pt}
\caption{\textbf{Semantic-memory staleness and age conditioning.} Success rate (SR) on RoboLab-120 with 10\,cm pose randomization (960 episodes per row). Ages are reported at action-expert calls; refresh counts are nominal. $\Delta$ reports percentage-point differences from fresh memory with task-bootstrap 95\% intervals. $^\dagger$Beyond the training age range. Protocol: Appendix~\ref{supp:memory_staleness}.}
\label{tab:memory_staleness}
\resizebox{\linewidth}{!}{%
\begin{tabular}{lcccc}
\toprule
Semantic memory & Age (steps) & Refreshes / chunk & SR (\%) & $\Delta$ (pp) \\
\midrule
Fresh memory & 0 & 1 & 20.5 & -- \\
Delayed, $N{=}8$ & 8 & 1 & \textbf{23.4} & $+2.9$ [$+0.2$, $+5.8$] \\
Delayed, $N{=}16$ & 16, 24 & 1/2 & 15.6 & $-4.9$ [$-8.5$, $-1.5$] \\
Delayed, $N{=}30^\dagger$ & 32--56 & 1/4 & 9.9 & $-10.6$ [$-14.7$, $-6.6$] \\
\midrule
Delayed, $N{=}16$, age hidden & 16, 24 & 1/2 & 11.2 & $-9.3$ [$-13.3$, $-5.3$] \\
\bottomrule
\end{tabular}%
}
\end{table}

\begin{table}[!t]
\centering
\small
\caption{\textbf{RTC on ARP's vision--action expert.} Success rate (SR) and execution metrics on an 18-task RoboLab subset. EE denotes end-effector; ISJ is integrated squared jerk and SPARC is spectral arc length. Arrows indicate better values. Time and smoothness are averaged over each policy's successful episodes.}
\label{tab:sync-vs-async}
\resizebox{\linewidth}{!}{%
\begin{tabular}{lcccc}
\toprule
Policy & SR (\%) $\uparrow$ & Time (s) $\downarrow$ & EE ISJ $\downarrow$ & EE SPARC $\uparrow$ \\
\midrule
ARP & 31.9 & 19.5 & 456 & -5.11 \\
ARP + RTC & 37.5 & 17.1 & 296 & -3.90 \\
\bottomrule
\end{tabular}%
}
\end{table}

\begin{table*}[!t]
\centering
\small
\caption{\textbf{Orchestration efficiency on RoboLab-120.} Usage and API-equivalent costs are per episode; $s$ scales planner latency. Plan age is the median update delay. $^\dagger$GPT-6 Astra~\cite{yang2026astranomy} uses a different simulator and success-only usage averages; $\uparrow297\times$ compares its reported cost with asynchronous $s=1$. Details are in Appendix~\ref{supp:orchestration_efficiency}.}
\label{tab:harness_efficiency}
\begin{tabular}{lcccccc}
\toprule
& Success & VLM calls & Planner compute & Tokens & API cost & Plan age \\
Configuration
& (\%) $\uparrow$
& / episode $\downarrow$
& (PFLOPs / episode) $\downarrow$
& (k / episode) $\downarrow$
& (\$ / episode) $\downarrow$
& (s) \\
\midrule
Synchronous
& 47.2 & 13.7 & $\sim$3.6 & 132 & 0.0221 & -- \\
Asynchronous ($s=1$)
& 43.9 & 9.2 & 0.98 & 35.0 & 0.0066 & 9.6 \\
Asynchronous ($s=2$)
& 41.4 & 5.8 & 0.61 & 22.1 & 0.0042 & 19.9 \\
Asynchronous ($s=4$)
& 35.5 & 4.0 & 0.43 & 15.2 & 0.0029 & 43.7 \\
\midrule
GPT-6 Astra as policy$^\dagger$
& 66.3 & 42.4 & -- & 160 & $1.96\;(\uparrow 297\times)$ & -- \\
\bottomrule
\end{tabular}
\end{table*}

\subsection{Failure Modes}

\begin{figure}
    \centering
    \includegraphics[width=\linewidth]{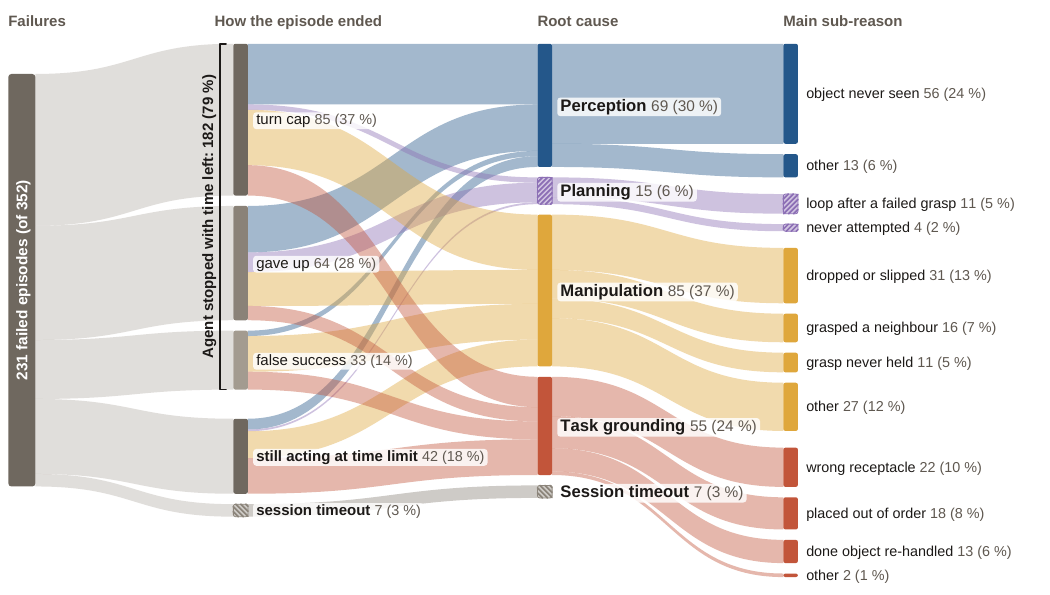}
    \caption{Failure categories and execution endings for 231 failures from the earlier RoboLab-Long synchronous cohort (34.4\% success), before the reruns reported in Table~\ref{tab:robolab_long_leaderboard_compact}.}

    \label{fig:failure_analysis}
\end{figure}

Figure~\ref{fig:failure_analysis} analyzes an earlier synchronous evaluation (121 successes and 231 failures in 352 episodes), before the rendering and infrastructure reruns used in Table~\ref{tab:robolab_long_leaderboard_compact}. It separates failure categories from execution endings; these counts do not describe the updated table cohort. Manipulation and perception account for the largest shares of failures, followed by task grounding. Typical execution failures include slips, drops, and grasping neighboring objects, while perception failures frequently involve targets that are never localized. Task-grounding errors include selecting the wrong receptacle, violating placement order, and manipulating objects whose subgoals have already been satisfied. These failures highlight the difficulty of maintaining consistent object identities and task progress across successive interactions.

Recovery and termination decisions often compound these initial errors. In 79\% of failed episodes, the agent stops before exhausting the simulation budget, either by reaching the planning-turn limit, giving up, or incorrectly declaring success. Repeated unsuccessful attempts can consume the planning budget without producing useful progress, while false success declarations prevent further recovery. The robot remains idle for 39\% of control steps across failed episodes. These observations suggest that improving long-horizon reliability requires both stronger perception and manipulation skills and better mechanisms for verifying outcomes, tracking completed subgoals, and changing strategy after failure.

\subsection{Ablations}
\label{sec:ablations}

\paragraph{Semantic-memory staleness and age conditioning.}
We evaluate one ARP checkpoint under fresh and delayed semantic-memory conditions on RoboLab-120, with object positions randomized by up to 10\,cm to test memory-age sensitivity across varied spatial configurations while preserving the task instructions. Tab.~\ref{tab:memory_staleness} reports success over eight episodes per task. Memory delays are emulated on the simulator clock; this study measures task performance rather than runtime savings. Scheduling details are in Appendix~\ref{supp:memory_staleness}.

At the nominal eight-step delay ($N=8$), ARP achieves 23.4\% success, compared with 20.5\% with fresh memory. Success falls as memory becomes older, reaching 9.9\% when all sampled ages exceed the training range ($N=30$). Providing the correct age helps at $N=16$: success is 15.6\%, versus 11.2\% when the same delayed memory is presented as current, a gain of 4.4 percentage points (task-bootstrap 95\% CI $[1.5,7.6]$). This supports the usefulness of memory-age conditioning at inference; it does not separately isolate the effect of training with stale memory.

\subsection{Action-Chunk Execution}
\label{sec:analysis_arp}

We apply real-time chunking (RTC) to ARP's vision--action expert, overlapping prediction with execution and conditioning each chunk on an action prefix. On an 18-task RoboLab subset, RTC improves success and end-effector smoothness (Table~\ref{tab:sync-vs-async}). Time and smoothness averages use each policy's successful episodes, limiting direct comparison. Evaluation details are in Appendix~\ref{supp:arp_rtc}.

\subsection{Asynchronous Orchestration Efficiency}
\label{sec:orchestration_efficiency}

On RoboLab-120, asynchronous ARP-Agent continues execution at a median plan age of 9.6\,s, achieving 43.9\% success versus 47.2\% for blocking execution (Table~\ref{tab:harness_efficiency}). Blocking pauses simulation during planning, so wall-clock budgets are not matched. Asynchronous execution uses 33\% fewer planner calls and $3.7\times$ less estimated planner computation. Savings reflect both planning frequency and stateless complete-plan prompts, compared with individual tool calls in a growing conversation.

GPT-6 Astra's reported API costs~\cite{yang2026astranomy} and our local planner's API-equivalent estimates cover different components and evaluation protocols; their ratio does not measure total-system savings. Appendix~\ref{supp:orchestration_efficiency} details timing and cost accounting.

Doubling planner latency reduces estimated planner computation by 38\% and success by 2.5 percentage points. At fourfold latency, success remains above standalone ARP despite a median plan age of 43.7\,s. Execution thus remains productive between delayed updates, although longer delays reduce task completion.

\section{Discussion and Conclusions}

We presented asynchronous reasoning and control at two levels. ARP combines fresh visual observations with stale semantic KV-cache memory; ARP-Agent reconciles delayed plan revisions with ongoing execution.

Across Franka tasks, dynamic-object grasping, RoboLab-120, and RoboLab-Long, ARP improves repeated object-transfer progress and robustness to target motion, while RTC improves action smoothness. Combining ARP with planning, perception, and geometric skills raises RoboLab-120 success from 25.2\% to 43.9\% and improves long-horizon task completion and subgoal progress. All ARP-Agent components run locally with publicly available model weights and public robot training data, without external model API calls. Asynchronous orchestration reduces estimated planner computation and token cost, with a measurable trade-off in task success relative to blocking execution.

Long-horizon reliability remains limited by perception errors, manipulation failures, and incomplete verification of task progress. Increasingly delayed planning updates further reduce completion, and our timing protocol accounts for planner delay rather than all system latencies. Future work should examine broader robot training data, more reliable outcome verification, and evaluation with complete hardware and wall-clock accounting.

\section*{LLM Usage Statement}

We used OpenAI Codex and Anthropic Claude to assist with manuscript drafting and revision, literature search and reference checks, technical discussion, code development and debugging, and preparation of analysis scripts, tables, and figures. This assistance affected text throughout the manuscript and supplement, supporting software, and the presentation and checking of experimental results. Generative models used as components of ARP and ARP-Agent are described in the method and experimental sections. The authors retain full responsibility for the scientific claims, citations, software, and reported results.

\section{ACKNOWLEDGMENT}
This research was partially funded by the Ministry of Education and Science of Bulgaria (support for INSAIT, part of the
Bulgarian National Roadmap for Research Infrastructure) and supported supported with computational resources provided by Google Cloud Platform (GCP).

{\small

}

\clearpage
\appendix
\addtocontents{toc}{\protect\setcounter{tocdepth}{2}}
\setcounter{figure}{0}
\setcounter{table}{0}
\setcounter{equation}{0}
\renewcommand{\thefigure}{S\arabic{figure}}
\renewcommand{\thetable}{S\arabic{table}}
\renewcommand{\theequation}{S\arabic{equation}}
\renewcommand{\theHfigure}{supp.\arabic{figure}}
\renewcommand{\theHtable}{supp.\arabic{table}}
\renewcommand{\theHequation}{supp.\arabic{equation}}
\makeatletter
\twocolumn[{%
\begin{minipage}{\textwidth}
\begin{center}
 {\Large\bfseries\thetitle\par}
 \medskip
 {\large Supplementary Material\par}
\end{center}
\bigskip
\pdfbookmark[0]{Contents}{supplement-contents}
\begingroup
\@twocolumnfalse
\tableofcontents
\endgroup
\end{minipage}
\bigskip
}]
\makeatother

\section{Timing and Computation}
\label{supp:orchestration_efficiency}
This section explains how planning delays enter the simulation clock and how we estimate planner computation and cost. It also describes the protocol differences needed to interpret the benchmark comparisons.
\subsection{Inference and Simulation Time}
\label{supp:inference_timing}
The policy timing study controls memory age rather than measuring VLM runtime. ARP's nominal semantic launch period and delay are both eight control steps. Memory age ranges from 8 to 15 steps between updates and is eight steps at regular action-expert requests. 
The first request uses current memory, and each agent handover resets the local policy clock. The server reconstructs memory from the selected older observation.
At the task level, ARP-Agent accounts for the delay against the simulation budget. GPT-Policy~\cite{cheng2026context} and Astranamy~\cite{yang2026astranomy} on the other hand pause the simulation during high-latency model API calls which seperates the simulation clock wrt to wall-clock. 


\begin{figure*}[!t]
 \centering
 \includegraphics[width=\linewidth]{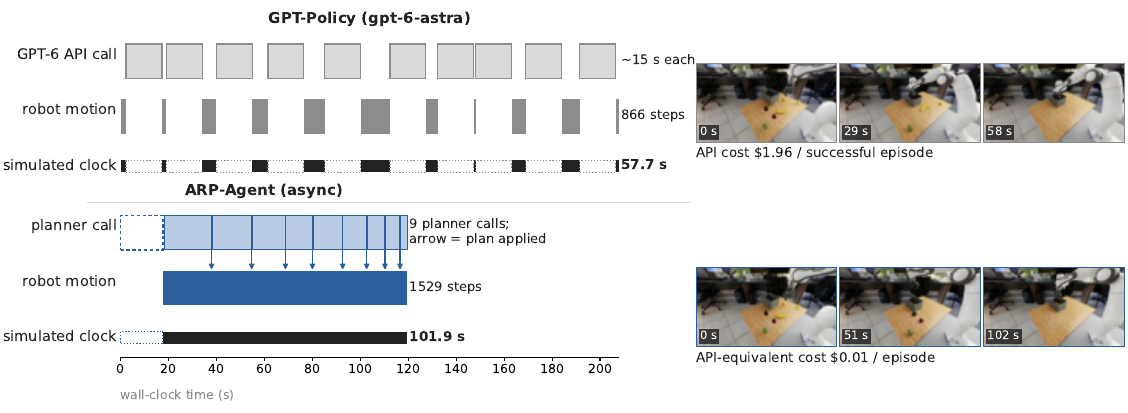}
 \caption{\textbf{Planning and execution use time differently.} \textit{Left}: GPT-Policy pauses simulation during inference; ARP-Agent overlaps planning and execution. GPT-Policy pauses are illustrative, while ARP-Agent calls are logged. \textit{Right}: episode frames and planner cost (GPT-6 Astra as reported for RoboLab-120~\cite{yang2026astranomy}; ours at API-equivalent token prices).}
 \label{fig:sim_clock_accounting}
\end{figure*}

\paragraph{Comparison with blocking GPT-Policy.}
We report a side-by-side comparison of ARP-agent vs GPT-Policy. 
In a six-object episode with matched initial layouts, illustrated in Fig.~\ref{fig:sim_clock_accounting}, both systems succeed: GPT-Policy uses 57.7\,s of simulation and ARP-Agent uses 101.9\,s. GPT-Policy's paused the simulator during model calls; ARP-Agent's clock includes planner delay after initialization, with execution continuing during that delay.

An estimated wall-clock comparison as well as token-cost comparison in Fig.~\ref{fig:sim_clock_accounting}, highlights the advantages of Asynchronous processes both in terms of compute as well timing efficiency. 


\subsection{Planner Computation and Cost}
The reported computation savings come from both fewer planner calls and shorter prompts. The blocking interface returns individual tool calls in a growing conversation. The asynchronous interface returns complete plans from stateless prompts. Together, these differences yield the reported $3.7\times$ reduction in estimated planner computation. The latency sweep holds the asynchronous interface fixed and delays plan updates, showing how less frequent planning trades computation for task success.

We estimate planner computation as $2NT_{\mathrm{computed}}$, where $N$ is the decoder parameter count (approximately 31.2 billion) and $T_{\mathrm{computed}}$ counts processed tokens, excluding input tokens served from the prefix cache. Generated-token computation includes the output head. This estimate covers the planner alone and omits attention-score operations and the vision tower.

API-equivalent cost expresses planner token usage at a common price; it is not the operating cost of our local system. We price input and output tokens at \$0.16 and \$0.64 per million, respectively, including cache hits and averaging over successful and failed episodes. Perception, action generation, simulation, and hardware rental are excluded. Longer RoboLab-Long tasks require more blocking planner turns, so the token-cost ratio differs between benchmarks.

The external GPT-6 Astra cost provides context, not a matched total-system comparison. It uses a different simulator, five episodes per task, and usage averaged over successful episodes. Our published baseline scores use the November 15, 2026 leaderboard snapshots~\cite{robolabLeaderboard2026,ladderLeaderboard2026}.

\subsection{Benchmark Protocol Differences}
\label{supp:benchmark_protocols}
Our RoboLab evaluations use an Isaac Sim 6 port with different arm gains and a different gripper actuator model from the public release. Published leaderboard scores therefore provide context rather than a controlled ranking against our results. Episode counts and scoring rules are given in Appendix~\ref{sec:franka_appendix}.

For RoboLab-Long, the agent variants use two passes over four starting layouts per task, while standalone ARP uses eight distinct official layouts. These evaluations have equal episode counts but do not share all initial states. Rendering and session-timeout failures were replaced independently of task outcome. The main paper's failure analysis describes an earlier synchronous cohort, before these replacements.

\section{ARP Architecture and Training}
\label{sec:appendix_arp}
ARP lets visual control respond to the current scene while a larger VLM updates semantic context. The following sections explain how the experts share memory, how training prepares them for delayed updates, and how the policy generates actions.

\subsection{Semantic Memory and the Vision--Action Expert}
The semantic memory lets the action expert reuse the VLM's interpretation of the task and scene. PaliGemma-3B encodes the instruction and camera images. We extract keys and values from eight transformer layers, then use layer normalization and linear projection to match the action expert's width. A learned positional embedding preserves token order. The expert projects this memory into attention keys and values, so it can read semantic context without another VLM forward pass. The projection changes feature width, while token count follows the VLM prefix length.

Fresh visual input gives the action expert a direct path for responding to scene changes. The expert is a Gemma-300M transformer with its own DINOv3 ViT-B/16 encoder. Current image features, projected proprioception, and action tokens attend to one another and to semantic memory. The images describe the scene now; the memory supplies task context from an earlier observation.

Temporal rotary embeddings tell the expert how old that context is. At control step $t$, let $\tau(t)$ be the observation step used to produce the latest available memory. Its age is $d_t=t-\tau(t)$. We place memory keys at $-d_t$ and current tokens at zero, using 16 temporal rotary pairs with wavelengths spanning 4--256 control steps. Age is applied when reading memory, so it can advance without changing the stored content.

\subsection{Training with Temporally Offset Observations}
Training teaches the action expert to use older semantic context alongside fresh observations. We train on DROID, initializing the semantic and visual encoders from pretrained PaliGemma and DINOv3 weights. Each semantic forward pass supplies memory for four action targets from the same demonstration: one at the semantic observation and three at later frames. Each target has its own images, proprioception, and action chunk. Later offsets are sampled from $\mathcal{U}\{1,\ldots,12\}$ with probability 0.8 and $\mathcal{U}\{13,\ldots,30\}$ otherwise.

This sampling both exposes the expert to delayed context and shares an expensive semantic forward pass across several action targets. Action-loss gradients stop at the semantic backbone boundary. We use a batch size of 2,048, warm up the learning rate to $10^{-4}$ over 1,000 steps, and apply cosine decay toward $5\times10^{-6}$ over a 60,000-step schedule.

\subsection{Action Generation}
The expert generates an action chunk by iteratively refining noise, conditioned on current observations and semantic memory. We train it with flow matching~\citep{lipman2024flow}. For a target chunk $a$ and Gaussian noise $\epsilon$, we form
\begin{equation}
 x_r=(1-r)a+r\epsilon,\qquad v^*=\epsilon-a.
 \label{eq:flow_interpolant}
\end{equation}
The conditioning $c_t=(I_t,s_t,M_{\tau(t)},d_t)$ contains current images $I_t$, robot state $s_t$, semantic memory $M_{\tau(t)}$, and its age $d_t$. The training objective is
\begin{equation}
 \mathcal{L}_{\mathrm{FM}}=\mathbb{E}\!\left[\|u_\theta(x_r,r;c_t)-v^*\|_2^2\right].
 \label{eq:flow_matching_loss}
\end{equation}
One flow time is sampled per target chunk as $r=0.001+0.999b$, where $b\sim\mathrm{Beta}(1.5,1)$. At inference, 10 Euler steps integrate from noise to a 16-step action chunk. The RoboLab executor applies eight actions at 15\,Hz before requesting another chunk.

Real-time chunking (RTC)~\cite{black2025training} lets the robot execute one chunk while the expert predicts its continuation. Actions already committed for execution form a prefix for the next prediction. These actions enter without noise, carry a prefix indicator, and are excluded from the loss. RTC therefore addresses action-generation delay alongside ARP's handling of delayed semantic memory.
\section{ARP-Agent Planning and Execution}
\label{supp:arp_agent}
ARP-Agent extends asynchronous computation to task planning: the robot executes available skills while a planner revises the remaining work. The modules share object descriptions, 3D geometry, and execution reports. This task-level memory is text and scene information, distinct from ARP's KV-cache-based semantic memory.

\subsection{From Observations to a Plan}
The planner returns several operations at once so the robot can continue between updates. We use a locally hosted Qwen3-VL-32B-Instruct model. It receives the task instruction, external and wrist images, object geometry, active and pending operations, and recent outcomes. It returns object queries and an ordered sequence of \texttt{pick}, \texttt{place}, and \texttt{place\_beside} operations, with optional interruption and completion flags.

The complete system prompt and a recorded observation message are provided in Appendix~\ref{supp:qwen_prompt}.

Perception grounds these operations in the current scene. SAM~3 masks and calibrated depth provide object positions, extents, orientations, and heights in the robot base frame. Planar hints help distinguish similar instances, and depth estimates the support surface. Perceptual checks track whether an object is held or has reached its destination. The planner uses these estimates and execution feedback, without access to simulator object poses or success labels.

\subsection{Combining Learned and Geometric Skills}
Geometric skills bring the gripper to the intended interaction; ARP handles grasping and release. The executor uses perceived geometry to approach, lift, and transport objects. Inverse kinematics converts Cartesian paths into joint targets with smooth waypoint timing. Near the object or destination, control passes to ARP, which receives current images, robot state, and the original task instruction. The executor selects the interaction through gripper positioning.

Execution checks determine when to continue or recover. A grasp handover ends when the grasp settles, motion stalls, the policy leaves the permitted region, or time runs out. A release handover ends when the gripper opens or another termination condition is reached. The executor then checks the gripper and perceived object state. Geometric grasps can handle clutter or recover from an unsuccessful policy attempt, allowing learned and geometric skills to support the same plan.

\subsection{Updating a Plan During Execution}
After obtaining an initial plan and the requested object detections, the agent overlaps planning with execution. A revision based on observation step $t_k$ becomes available at $t_k+\lceil f s\lambda_k\rceil$, where $f=15$\,Hz and $\lambda_k$ is measured planner latency. The efficiency study scales this delay by $s$. Once the revision is available, the next request uses a fresh observation.

A late plan must account for work completed while the planner was running. The executor matches returned operations against recorded completions using object descriptions, destinations, relations, and instance hints. It removes the prefix through the latest matched completion, then drops leading operations incompatible with the current gripper state. This assumes the matched prefix has been resolved. The remaining sequence replaces pending work, while the active operation normally continues.

\begin{figure*}[t]
 \centering
 \includegraphics[width=\linewidth]{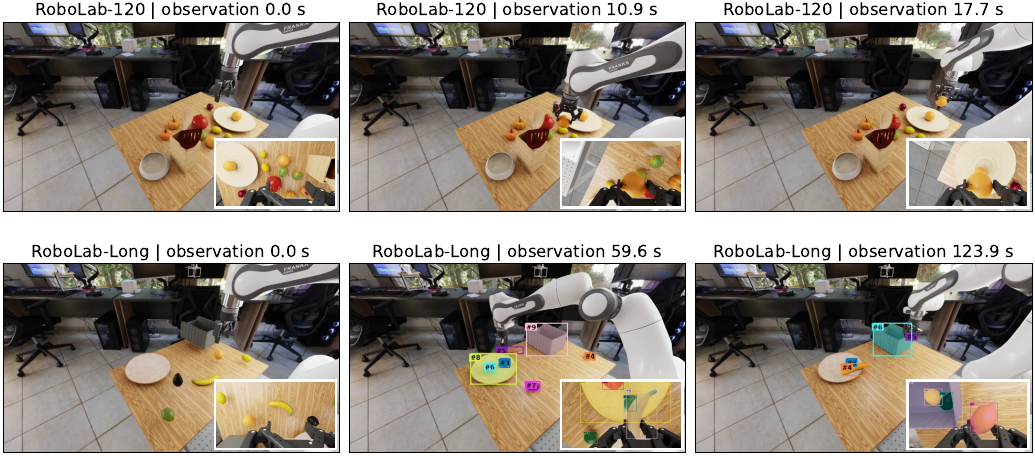}
 \caption{\textbf{Planner observations during asynchronous execution.} The scene changes as the robot places oranges on a plate (top, RoboLab-120) and routes objects between a plate and a bin (bottom, RoboLab-Long). Each row shows early, middle, and late planner observations with wrist views inset; boxes show recorded perception outputs. Times mark observation acquisition in simulation.}
 \label{fig:supp_robolab_observations}
\end{figure*}

Interruptions are restricted to avoid repeatedly stopping the same interaction. The planner must have observed the scene at least three seconds after the active operation began, and may interrupt once per object. Only undispatched commands are cancelled. The robot holds position if no operation is available, and accepts a completion declaration only when its gripper is empty. Figure~\ref{fig:supp_robolab_observations} shows scenes observed during these updates.

\section{Evaluation Details}
\label{sec:franka_appendix}
\subsection{Real-World Franka Tasks}
We evaluate real-world manipulation on three Franka tasks using the DROID hardware setup and an NVIDIA A6000 for inference. Each policy is evaluated over ten trials per task with a fixed 90\,s budget, allowing us to compare the amount of manipulation completed within the same allotted time. We score progress from the final scene according to the task-specific criteria below. These experiments evaluate the standalone policies; ARP-Agent is evaluated in simulation.

\paragraph{Vegetables in bowl.}
The robot must place five toy vegetables into a bowl. Each correctly placed vegetable contributes $1/5$ to the score.

\paragraph{Cup stacking.}
The robot must nest four cups into a single stack. Each additional correctly nested cup contributes $1/3$ to the score.

\paragraph{Markers in bowl.}
The robot must place three markers into a bowl. Each correctly placed marker contributes $1/3$ to the score.

On the vegetables-in-bowl task, ARP achieves approximately $2.4\times$ the mean progress of $\pi_{0.5}$, averaged over all trials. This comparison measures task progress under the fixed time budget.

\subsection{RoboLab-120}
\label{sec:robolab}
We use RoboLab-120 to evaluate performance across diverse objects, instructions, and spatial requirements. We run ten seeded episodes for each of 120 tasks, totaling 1,200 episodes per configuration. The simple, moderate, and complex tiers contain 64, 39, and 17 tasks. We use nominal initial poses and default instructions, and average binary episode success with equal weight per task.

The comparison between standalone ARP and ARP-Agent evaluates the contribution of the complete agent system. Standalone ARP receives RGB images and proprioception, whereas ARP-Agent additionally uses calibrated depth, perception, planning, and geometric skills. Consequently, this comparison does not isolate the effect of planning alone.

\subsection{RoboLab-Long}
\label{supp:robolab_long_protocol}
We use RoboLab-Long to evaluate the completion of extended manipulation sequences. Its 44 tasks require one to ten subgoals, including split destinations, ordered placement, repeated objects, transfers, relays, and revisits. We run eight episodes per task, totaling 352 per configuration. Task-specific budgets range from 60 to 330\,s of simulation at 15\,Hz. The agent receives both RGB views and calibrated depth from the external camera.

We report both full-task success and subgoal completion to distinguish complete solutions from partial progress. Full-task success requires all final-state and ordering conditions to hold, whereas subgoal completion measures the fraction of required subgoals satisfied across episodes. LADDER credits a relay or revisit leg only after the earlier legs for that object are completed in order. These evaluation rules are independent of the agent's perceptual checks.


\subsection{Dynamic-Object Grasping}
\label{supp:dyn_bump_details}
We evaluate adaptation to unexpected target motion using a dynamic-object grasping task. A Franka Panda grasps a red ball, blue ball, or green cylinder among two static distractors. The target starts with a random velocity toward the robot and slows under friction.

To control for training data, both policies use the same synthetic dataset: 2,500 successful trajectories per task from a scripted interception controller with a 0.23\,s prediction horizon. Training and evaluation seeds are separate. We compare smaller InternVL3.5-1B variants of ARP and the $\pi_0$-style baseline under this shared training setup. These differ from the PaliGemma/DROID policies used on Franka and RoboLab.

The robot runs an operational-space pose controller at 20\,Hz. Observations contain images, end-effector position and axis-angle orientation, and gripper state. The seven-dimensional action specifies Cartesian and axis-angle displacements plus a gripper command.

To test adaptation after reaching has begun, we perturb the target velocity during execution. Each episode schedules at most one velocity change with probability $p\in\{0.5,1\}$, between steps 30 and 100 (1.5--5.0\,s). We add a planar velocity of 0.20--0.45\,m/s within $\pm100^\circ$ of the initial travel direction. The perturbation is skipped if the target is already more than 4\,cm above the table, so $p$ is a scheduling probability. Policies share seeds for initial conditions and scheduled perturbations.

\begin{figure*}[t]
    \centering
    \includegraphics[width=0.95\linewidth]{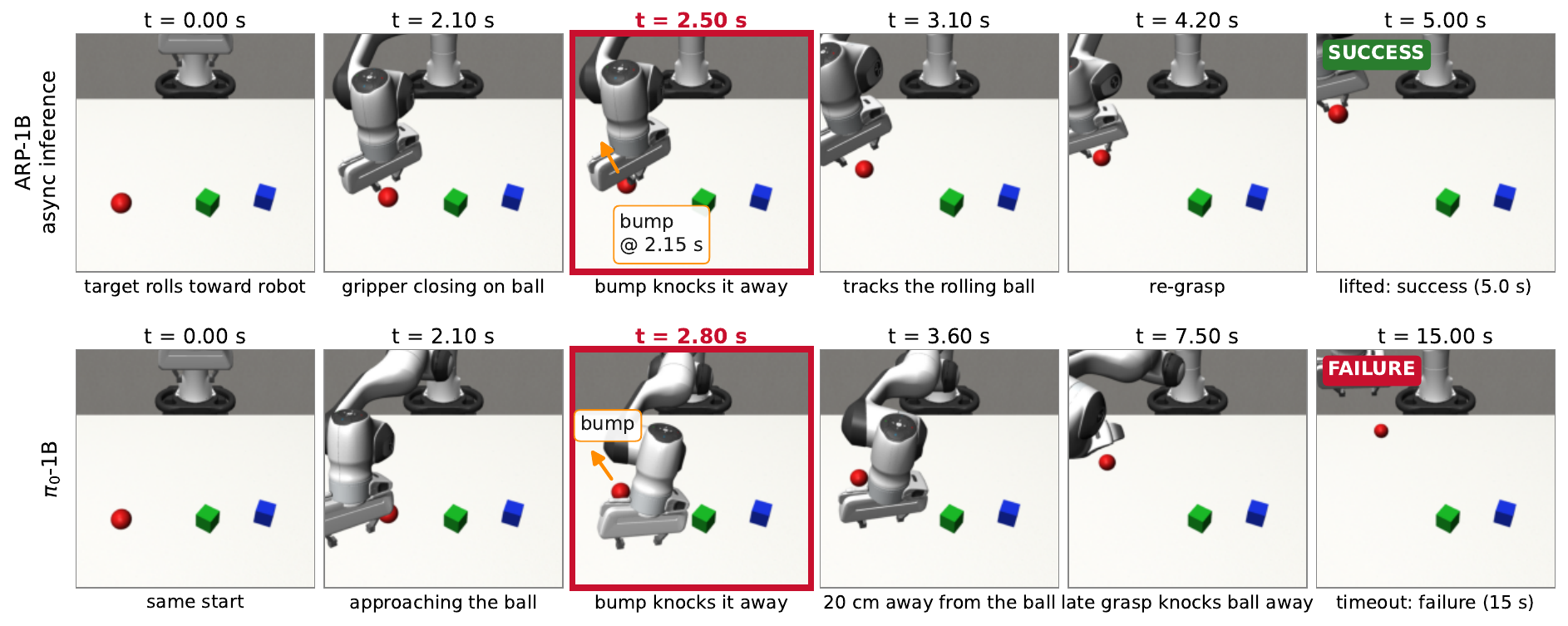}
    \caption{\textbf{Reaction to unexpected target motion.} After a velocity perturbation, ARP adjusts its reach and grasps the moving ball, while $\pi_0$ misses the target in this example.}
    \label{fig:dynamic_bump}
\end{figure*}

Success requires grasping the target and holding it more than 6\,cm above the table for 10 consecutive steps. An episode ends on success, when the target falls off the table, or after 300 steps (15\,s). Figure~\ref{fig:dynamic_bump} illustrates how the two policies respond to a change in target motion.

\subsection{Semantic-Memory Staleness}
\label{supp:memory_staleness}
We ablate semantic-memory staleness to assess how delayed context affects task success and whether explicit age conditioning mitigates this effect. The same trained policy receives current or delayed semantic observations, while its action expert receives fresh observations every eight control steps. We evaluate eight episodes per RoboLab-120 task, with object positions randomized by up to 10\,cm, for 960 episodes per condition.

We vary the memory-update interval and delivery delay jointly. Fresh memory is encoded before each action chunk. In delayed conditions, an update is scheduled after at least $N$ steps and becomes available $N$ steps later. On the eight-step request grid, the expert reads ages of 8 steps for $N=8$; 16 or 24 for $N=16$; and 32, 40, 48, or 56 for $N=30$. The last condition exceeds the training age range. To isolate the contribution of age information at inference, the age-hidden condition uses the same older observation as $N=16$ but reports age zero.

At $N=16$, providing the correct memory age increases observed success from 11.2\% to 15.6\%. This result supports explicit age conditioning, although success still declines as semantic memory becomes older. We compute 95\% confidence intervals from 4,000 task-bootstrap resamples, preserving episodes within each task.

This ablation measures tolerance to temporal mismatch: the server reconstructs memory from stored observations, so the experiment does not measure concurrent-inference speed. The nominal-delay condition uses an earlier serving revision, while the remaining conditions share a revision. In addition, the randomized object poses distinguish this study from the nominal-pose benchmark evaluation.

\subsection{Real-Time Chunking}
\label{supp:arp_rtc}
We evaluate real-time chunking (RTC) to assess whether overlapping action prediction and execution preserves motion smoothness. We evaluate 18 simple RoboLab tasks whose objects and goals are represented in DROID, estimated by matching object nouns and goal phrases, including synonyms, in demonstration instructions. Both modes use the same prefix-trained policy and eight layouts per task, for 144 episodes per configuration.

The two execution modes differ in whether action prediction is conditioned on the committed action prefix. With RTC, each prediction receives the eight-step prefix already scheduled for execution; the baseline executes without prefix conditioning. We evaluate this comparison separately from the main benchmark, where ARP runs without RTC in comparisons with methods such as VoLo~\cite{chen2026volo} and $\pi_{0.5}$~\cite{pi05}.

We measure success over all episodes and average execution time and smoothness over each policy's successful episodes. Smoothness is quantified by integrated squared jerk (ISJ; lower is smoother) and spectral arc length (SPARC; closer to zero is smoother). ARP with RTC produces smoother actions and a slightly higher observed success rate on this subset. Because the sets of successful episodes can differ between modes, the time and smoothness averages describe their respective successful executions rather than a paired comparison.


\section{Additional Analysis}
\subsection{Task Difficulty and Long-Horizon Progress}
Both ARP-Agent variants improve on standalone ARP in every RoboLab-120 difficulty tier (Figure~\ref{fig:supp_benchmark_analysis}a). Compared with blocking execution, asynchronous execution succeeds less often on simple tasks and slightly more often on moderate and complex tasks.

On RoboLab-Long, we score ordered subgoals at each quarter of the simulation budget, retaining credit after success. Asynchronous ARP-Agent reaches 59.1\% completion by the third quarter and 62.1\% by the end, close to the blocking agent's 64.3\% (Figure~\ref{fig:supp_benchmark_analysis}b). The gap is larger for full-task success (28.4\% versus 36.1\%).

\noindent\begin{minipage}{\linewidth}
 \centering
 \includegraphics[width=0.82\linewidth,trim=0 0 260 0,clip]{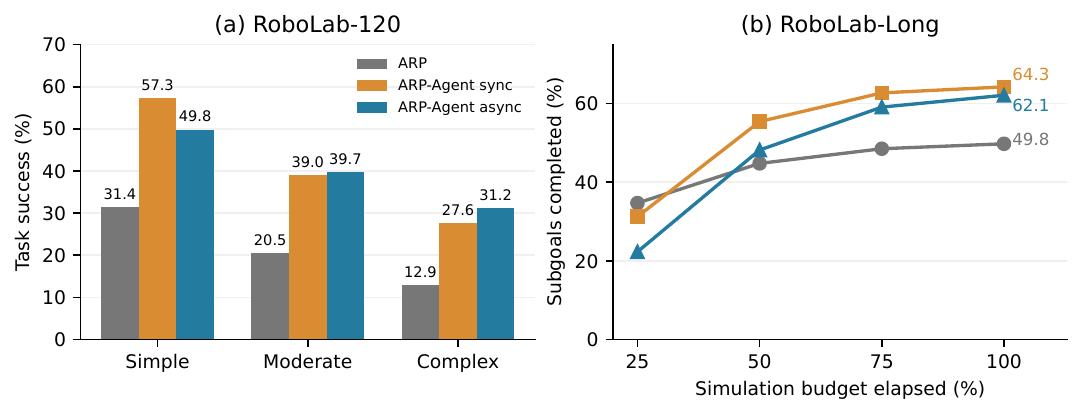}\par
 \includegraphics[width=0.82\linewidth,trim=260 0 0 0,clip]{figures/supp_benchmark_analysis.pdf}
 \captionof{figure}{\textbf{Benchmark performance and subgoal progress.} (a) RoboLab-120 task success across 640 simple, 390 moderate, and 170 complex episodes per policy. (b) RoboLab-Long subgoal completion over 352 episodes per policy.}
 \label{fig:supp_benchmark_analysis}
\end{minipage}\par

\onecolumn
\renewcommand{\thelstlisting}{S\arabic{lstlisting}}
\section{Qwen Planner Prompt}
\label{supp:qwen_prompt}
We document the planner interface by reproducing the complete system prompt and a recorded observation message from asynchronous ARP-Agent. Each call consists of the system prompt and a user message containing the current observation. Since calls do not retain conversation history, the observation includes the active plan, recent outcomes, and task progress needed for replanning. The same interface is used to refine the initial plan once object detections become available.

\subsection{System Prompt}
\label{supp:qwen_system_prompt}
Listing~\ref{lst:qwen_system} reproduces the system prompt generated for the RoboLab embodiment by the evaluated prompt builder, preserving its wording and JSON example. The table height is populated from the episode's depth estimate and is 0.004\,m in this example. Although the prompt retains the shared tool instructions sent to the model, the asynchronous caller supplies no tool schemas; it interprets the returned JSON as a plan for the executor.

\begin{lstlisting}[style=plannerprompt,caption={Complete asynchronous planner system prompt.},label={lst:qwen_system}]
You are controlling a robot through tool calls. Robot: Franka Panda (DROID platform in simulation, one 7-DoF arm, Robotiq 2F-85 parallel-jaw gripper).

WORLD FRAME. Right-handed, z up, metres. The robot base frame: origin at the robot base, +x forward away from the robot (toward the table centre), +y to the robot's left, +z up. The workspace is the table in front of the robot. The table surface is at z = 0.004 m.
All positions you read (object readouts, measured tool poses) and all positions you command are in this one frame,
so a measured object centroid can be used directly as a move_to target.

ARMS
- 'arm' arm: base at [0.0, 0.0, 0.0] m; TCP workspace x in [0.15, 0.85], y in [-0.60, 0.60], z in [-0.01, 0.65] m; tilt in [0, 100] deg; parallel-jaw gripper, max opening 85 mm. 7-DoF Franka arm with a Robotiq 2F-85 parallel-jaw gripper (85 mm stroke). The TCP is between the fingertips; the fingers are ~5 cm long, so a grasp at height z encloses roughly z..z+0.04 m of the object.

CAMERAS (one downscaled RGB image per camera every turn, annotated with masks and numbered boxes for the current queries)
- 'over_shoulder_left': fixed in the world, image 640x360. Fixed camera over the robot's left shoulder looking down at the table (front-left view of the scene).
- 'wrist': mounted on the arm arm's wrist (moves with it), image 512x288. Wrist camera on the gripper, looking along the fingers (the fingertips are at the bottom of the image).

ORIENTATION of the tool center point (TCP), three angles in degrees, each measured counter-clockwise from world +x seen from above:
- slot_yaw: heading of the jaw slot, i.e. the axis a held object lies along. For a top-down grasp of an elongated object, set slot_yaw to the object's major_axis_yaw so the jaws close across its narrow side.
  The jaws are symmetric, so slot_yaw and slot_yaw+180 give the same grasp; use values in [0, 180). The measured pose is reported that way too.
- tilt: angle of the tool axis away from straight down. 0 = pointing straight down (top-down grasp), 90 = horizontal (side grasp).
- tilt_direction: compass direction the fingertips tip toward. Has no meaning when tilt = 0.

PERCEPTION. A segmentation model (SAM) runs continuously on every camera on the object descriptions you set with change_query.
Each detection is reported per camera and is NOT fused across cameras: you decide which detections are the same physical object.
Readout fields: surface_centroid = mean of the visible surface points; centroid = estimated volumetric centre (surface xy, z halfway between bottom_z and top_z);
top_z / bottom_z / height in metres; major_axis_yaw = heading of the longest horizontal axis in degrees; extent_major / extent_minor = footprint size in mm;
conf = detector confidence; points = number of valid depth points. Re-query (change_query) to find a missed object or a part of one (e.g. "cap of the tube").
Wrist cameras see objects close up after you move above them; their readouts are usually the most accurate for grasping.

TOOLS. move_to (absolute TCP pose), set_gripper (0 closed .. 1 open),
change_query (replace the perception queries), update_todo (replace your short plan; it is repeated in every observation), done.
There are no relative moves, grasp macros or approach primitives: you plan the approach, descent, grasp, lift, transport and release yourself.
A move_to that violates the declared limits is refused (not clipped) and the reason is returned; choose a different pose.
The controller moves the TCP on a straight line and reports: reached, short, incomplete, not attempted or unknown, with the cause and the measured errors.
Gripper feedback tells you whether the jaws stalled on an object and at what aperture; use it and the wrist images to verify grasps.

RULES. Make exactly one tool call per turn. Think briefly before each call. Start by writing a plan (update_todo) and setting perception queries.
Verify outcomes from the readouts (e.g. top_z rising after a lift) rather than assuming success. Call done with outcome="success" only when the task is visibly complete,
and done with outcome="failure" if it cannot be completed. You have a limited step budget shown in every observation.

ASYNCHRONOUS OPERATION. You are the planner of a robot that NEVER stops to wait for you. An executor runs your plan item by item with learned and scripted skills (approach, grasp, carry, place) while you are thinking; by the time your answer arrives, the scene has moved on by the time you took. Every observation you get is the freshest one available, with the executor's progress reports since your last answer. Your job each time: look at the images and readouts, compare with the reports, and return the plan the executor should follow FROM NOW ON (items not started yet are replaced by yours; the item being executed continues unless you set preempt).

OUTPUT: answer with ONE JSON object and nothing else:
{
 "queries": [
  "colour + noun of every object the task involves (targets and destinations)"
 ],
 "plan": [
  {
   "do": "pick",
   "obj": "green lizard toy",
   "hint_xy": [
    0.45,
    0.1
   ]
  },
  {
   "do": "place",
   "obj": "green lizard toy",
   "rel": "in",
   "dest": "grey bin"
  },
  {
   "do": "place_beside",
   "obj": "red cube",
   "rel": "behind",
   "dest": "white bowl"
  }
 ],
 "preempt": false,
 "done": false,
 "note": "one sentence on what you see and why this plan"
}
- 'queries': object descriptions for perception (colour + noun, e.g. 'grey bin', never a bare noun).
- 'plan': the remaining steps in order. 'pick' grasps an object (optional hint_xy = which instance, robot frame metres). 'place' puts the HELD object in/on a container or support. 'place_beside' puts the held object beside a reference ('behind', 'in_front_of', 'left_of', 'right_of') -- use it to take an object OUT of a container (dest = the container) or to put it on the table next to something; 'place' with rel 'in'/'on' puts the object INTO/ONTO dest. One object at a time: pick, then place. Objects already in their destination need no items; an object already held needs no pick.
- 'preempt': true only if the current action is clearly wrong (wrong object, wrong destination) and must be stopped now. Your observation may be a few seconds old: a pick that is 'executing now' is usually fine -- do not interrupt it just because the object is not yet in the gripper.
- To put an object down on the table (take-out tasks), use 'place' with rel 'on' and dest 'table'.
- 'done': true only when EVERY goal object is verified in its destination by the readouts and nothing is held; then 'plan' is empty.
- Use the reports: a failed pick usually means another instance (hint_xy) or a different description; do not repeat an identical failed item more than once. When a report says no 'X' found, the NEXT item for that object MUST describe it differently: the colour you actually SEE on its body or lid (the task's words may not match what the cameras see), a different noun (container, cup, tub, packet, bottle, box), or a hint_xy on it.
- CONTAINMENT lines tell you which objects already sit inside which containers: never pick an object that is already in its destination; use them to decide 'done'.

SPATIAL WORDS (robot frame: x points away from the robot, y to its left): 'behind X' = farther from the robot than X (larger x, about the same y); 'in front of X' = closer to the robot (smaller x); 'left of X' = larger y; 'right of X' = smaller y. 'On / onto X' = resting on X's top surface, inside its footprint. 'In / into' a container = inside its walls, below the rim. For beside-relations put the object down on the table next to X and release it.

BOWLS, PLATES, TRAYS: grasp the rim, not the centre. Pick a rim point p = centre + (radius - 0.015 m) * u for a horizontal unit vector u (the side facing the robot is usually free), move above p with slot_yaw = heading(u) + 90 degrees so the jaws close across the wall, descend until the fingertips are ~0.025 m below the rim top (top_z), close, lift slowly. To stack a bowl into another, carry it by the rim, centre it over the other bowl (servo 'above'), lower, release.
\end{lstlisting}

\subsection{Observation Message}
\label{supp:qwen_observation}
Listing~\ref{lst:qwen_observation} gives the complete text from the third planner call in the oranges-to-plate rollout shown in Figure~\ref{fig:supp_robolab_observations}. The observation was acquired at simulation step 163. It records a placement in progress, the remaining plan, object measurements, and the latest execution report. These fields are regenerated on each call to provide the current execution context, with failed-operation summaries, missing-object notices, and perception warnings included when applicable. The reported turn budget counts planner calls rather than robot control steps.

The user message contains this text, then the label \texttt{[camera: over\_shoulder\_left]} and its JPEG image, followed by \texttt{[camera: wrist]} and its JPEG image. These are the external and wrist views of the middle observation in the figure's top row. Images are supplied as multimodal inputs, with available perception overlays, rather than converted to text. The measured readouts and executor reports below are reproduced without correction.

\begin{lstlisting}[style=plannerprompt,caption={Recorded user-message text for a planner update.},label={lst:qwen_observation}]
TASK: Put all the oranges on the plate
Turn 3/400; 397 steps remaining.
TODO: (empty; write one with update_todo)
Arms (measured):
  arm: TCP x=0.336 y=0.029 z=0.203 slot_yaw=120, tilt=0, tilt_direction=18; jaw opening 0.76 (64.3 mm)
Perception (query generation 2; queries=['orange fruit', 'white plate']). Detections are per camera and are NOT fused across cameras:
  [over_shoulder_left]
    "orange fruit"#1 conf=0.914 surface_centroid=<0.331,0.037,0.213> centroid=<0.331,0.037,0.119> top_z=0.235 bottom_z=0.004 height=0.231 major_axis_yaw=30deg extent_major=62mm extent_minor=42mm points=495
    "orange fruit"#2 conf=0.910 surface_centroid=<0.370,-0.314,0.064> centroid=<0.370,-0.314,0.059> top_z=0.086 bottom_z=0.032 height=0.055 major_axis_yaw=54deg extent_major=49mm extent_minor=40mm points=199
    "white plate"#3 conf=0.507 surface_centroid=<0.342,-0.308,0.030> centroid=<0.342,-0.308,0.025> top_z=0.046 bottom_z=0.004 height=0.042 major_axis_yaw=88deg extent_major=278mm extent_minor=253mm points=1662
  [wrist]
    "orange fruit"#4 conf=0.955 surface_centroid=<0.331,0.032,0.222> centroid=<0.331,0.032,0.119> top_z=0.235 bottom_z=0.004 height=0.231 major_axis_yaw=20deg extent_major=55mm extent_minor=42mm points=14043
GRIPPER NOW: HOLDING something (64 mm between the jaws).

PLAN STATE (version 2): executing now: {"do": "place", "obj": "orange fruit", "dest": "white plate", "rel": "on"} (started at step 152, now step 163; preempt only if you see it going wrong); queued: [{"do": "place", "obj": "orange fruit", "dest": "white plate", "rel": "on"}, {"do": "pick", "obj": "orange fruit", "hint_xy": [0.38, -0.32]}, {"do": "place", "obj": "orange fruit", "dest": "white plate", "rel": "on"}]
CONTAINMENT (from the 3D readouts): orange fruit at (0.33, 0.03) is on the table, outside every container; orange fruit at (0.37, -0.31) is INSIDE the white plate
EXECUTOR REPORTS (latest last):
  [step 152] pick orange fruit: Other objects lie within 10 cm of the orange fruit (~21 cm2), so the scripted grasp was used instead of ARP: grasp 'orange fruit': centred, slot yaw 120 deg, closed at z=0.042, lifted 12 cm -> the orange fruit IS in the gripper (64 mm, bottom_z=0.142).
\end{lstlisting}


\begin{thebibliography}{59}
\providecommand{\natexlab}[1]{#1}
\providecommand{\url}[1]{\texttt{#1}}
\expandafter\ifx\csname urlstyle\endcsname\relax
  \providecommand{\doi}[1]{doi: #1}\else
  \providecommand{\doi}{doi: \begingroup \urlstyle{rm}\Url}\fi

\bibitem[Ahn et~al.(2022)Ahn, Brohan, Brown, Chebotar, Cortes, David, Finn, Fu, Gopalakrishnan, Hausman, Herzog, Ho, Hsu, Ibarz, Ichter, Irpan, Jang, Ruano, Jeffrey, Jesmonth, Joshi, Julian, Kalashnikov, Kuang, Lee, Levine, Lu, Luu, Parada, Pastor, Quiambao, Rao, Rettinghouse, Reyes, Sermanet, Sievers, Tan, Toshev, Vanhoucke, Xia, Xiao, Xu, Xu, Yan, and Zeng]{saycan2022arxiv}
Michael Ahn, Anthony Brohan, Noah Brown, Yevgen Chebotar, Omar Cortes, Byron David, Chelsea Finn, Chuyuan Fu, Keerthana Gopalakrishnan, Karol Hausman, Alex Herzog, Daniel Ho, Jasmine Hsu, Julian Ibarz, Brian Ichter, Alex Irpan, Eric Jang, Rosario~Jauregui Ruano, Kyle Jeffrey, Sally Jesmonth, Nikhil Joshi, Ryan Julian, Dmitry Kalashnikov, Yuheng Kuang, Kuang-Huei Lee, Sergey Levine, Yao Lu, Linda Luu, Carolina Parada, Peter Pastor, Jornell Quiambao, Kanishka Rao, Jarek Rettinghouse, Diego Reyes, Pierre Sermanet, Nicolas Sievers, Clayton Tan, Alexander Toshev, Vincent Vanhoucke, Fei Xia, Ted Xiao, Peng Xu, Sichun Xu, Mengyuan Yan, and Andy Zeng.
\newblock Do as i can and not as i say: Grounding language in robotic affordances.
\newblock \emph{arXiv preprint arXiv:2204.01691}, 2022.

\bibitem[Assran et~al.(2025)Assran, Bardes, Fan, Garrido, Howes, Muckley, Rizvi, Roberts, Sinha, Zholus, et~al.]{assran2025v}
Mido Assran, Adrien Bardes, David Fan, Quentin Garrido, Russell Howes, Matthew Muckley, Ammar Rizvi, Claire Roberts, Koustuv Sinha, Artem Zholus, et~al.
\newblock V-jepa 2: Self-supervised video models enable understanding, prediction and planning.
\newblock \emph{arXiv preprint arXiv:2506.09985}, 2025.

\bibitem[Bai et~al.(2025)]{qwen3vl2025}
Shuai Bai et~al.
\newblock {Qwen3-VL} technical report.
\newblock \emph{arXiv preprint arXiv:2511.21631}, 2025.

\bibitem[Beyer et~al.(2024)Beyer, Steiner, Pinto, Kolesnikov, Wang, Salz, Neumann, Alabdulmohsin, Tschannen, Bugliarello, Unterthiner, Keysers, Koppula, Liu, Grycner, Gritsenko, Houlsby, Kumar, Rong, Eisenschlos, Kabra, Bauer, Bošnjak, Chen, Minderer, Voigtlaender, Bica, Balazevic, Puigcerver, Papalampidi, Henaff, Xiong, Soricut, Harmsen, and Zhai]{beyer2024paligemma}
Lucas Beyer, Andreas Steiner, André~Susano Pinto, Alexander Kolesnikov, Xiao Wang, Daniel Salz, Maxim Neumann, Ibrahim Alabdulmohsin, Michael Tschannen, Emanuele Bugliarello, Thomas Unterthiner, Daniel Keysers, Skanda Koppula, Fangyu Liu, Adam Grycner, Alexey Gritsenko, Neil Houlsby, Manoj Kumar, Keran Rong, Julian Eisenschlos, Rishabh Kabra, Matthias Bauer, Matko Bošnjak, Xi Chen, Matthias Minderer, Paul Voigtlaender, Ioana Bica, Ivana Balazevic, Joan Puigcerver, Pinelopi Papalampidi, Olivier Henaff, Xi Xiong, Radu Soricut, Jeremiah Harmsen, and Xiaohua Zhai.
\newblock {PaliGemma: A versatile 3B VLM for transfer}.
\newblock \emph{arXiv preprint arXiv:2407.07726}, 2024.

\bibitem[Black et~al.(2024)Black, Brown, Driess, Esmail, Equi, Finn, Fusai, Groom, Hausman, Ichter, et~al.]{black2024pi_0}
Kevin Black, Noah Brown, Danny Driess, Adnan Esmail, Michael Equi, Chelsea Finn, Niccolo Fusai, Lachy Groom, Karol Hausman, Brian Ichter, et~al.
\newblock $pi\_0$: A vision-language-action flow model for general robot control.
\newblock \emph{arXiv preprint arXiv:2410.24164}, 2024.

\bibitem[Black et~al.(2025{\natexlab{a}})Black, Brown, Darpinian, Dhabalia, Driess, Esmail, Equi, Finn, Fusai, Galliker, Ghosh, Groom, Hausman, brian ichter, Jakubczak, Jones, Ke, LeBlanc, Levine, Li-Bell, Mothukuri, Nair, Pertsch, Ren, Shi, Smith, Springenberg, Stachowicz, Tanner, Vuong, Walke, Walling, Wang, Yu, and Zhilinsky]{pi05}
Kevin Black, Noah Brown, James Darpinian, Karan Dhabalia, Danny Driess, Adnan Esmail, Michael~Robert Equi, Chelsea Finn, Niccolo Fusai, Manuel~Y. Galliker, Dibya Ghosh, Lachy Groom, Karol Hausman, brian ichter, Szymon Jakubczak, Tim Jones, Liyiming Ke, Devin LeBlanc, Sergey Levine, Adrian Li-Bell, Mohith Mothukuri, Suraj Nair, Karl Pertsch, Allen~Z. Ren, Lucy~Xiaoyang Shi, Laura Smith, Jost~Tobias Springenberg, Kyle Stachowicz, James Tanner, Quan Vuong, Homer Walke, Anna Walling, Haohuan Wang, Lili Yu, and Ury Zhilinsky.
\newblock $\pi_{0.5}$: a vision-language-action model with open-world generalization.
\newblock In \emph{9th Annual Conference on Robot Learning}, 2025{\natexlab{a}}.

\bibitem[Black et~al.(2025{\natexlab{b}})Black, Galliker, and Levine]{black2025real}
Kevin Black, Manuel~Y Galliker, and Sergey Levine.
\newblock Real-time execution of action chunking flow policies.
\newblock \emph{arXiv preprint arXiv:2506.07339}, 2025{\natexlab{b}}.

\bibitem[Black et~al.(2025{\natexlab{c}})Black, Ren, Equi, and Levine]{black2025training}
Kevin Black, Allen~Z Ren, Michael Equi, and Sergey Levine.
\newblock Training-time action conditioning for efficient real-time chunking.
\newblock \emph{arXiv preprint arXiv:2512.05964}, 2025{\natexlab{c}}.

\bibitem[Carion et~al.(2026)Carion, Gustafson, Hu, Debnath, Hu, Coll-Vinent, Ryali, Alwala, Khedr, Huang, Lei, Ma, Guo, Kalla, Marks, Greer, Wang, Sun, R{\"a}dle, Afouras, Mavroudi, Xu, Wu, Zhou, Momeni, HAZRA, Ding, Vaze, Porcher, Li, Li, Kamath, Cheng, Dollar, Ravi, Saenko, Zhang, and Feichtenhofer]{carion2026sam}
Nicolas Carion, Laura Gustafson, Yuan-Ting Hu, Shoubhik Debnath, Ronghang Hu, Didac~Suris Coll-Vinent, Chaitanya Ryali, Kalyan~Vasudev Alwala, Haitham Khedr, Andrew Huang, Jie Lei, Tengyu Ma, Baishan Guo, Arpit Kalla, Markus Marks, Joseph Greer, Meng Wang, Peize Sun, Roman R{\"a}dle, Triantafyllos Afouras, Effrosyni Mavroudi, Katherine Xu, Tsung-Han Wu, Yu Zhou, Liliane Momeni, RISHI HAZRA, Shuangrui Ding, Sagar Vaze, Francois Porcher, Feng Li, Siyuan Li, Aishwarya Kamath, Ho~Kei Cheng, Piotr Dollar, Nikhila Ravi, Kate Saenko, Pengchuan Zhang, and Christoph Feichtenhofer.
\newblock {SAM} 3: Segment anything with concepts.
\newblock In \emph{The Fourteenth International Conference on Learning Representations}, 2026.

\bibitem[Chen et~al.(2026{\natexlab{a}})Chen, Bagherieh, Appiah~Balaji, Gonzalez, Heravi, Islam, Li, Regli, Ren, {\v{S}}vec, Yoon, M{\"u}ller, and Koltun]{2026amp}
Chen Chen, Omid Bagherieh, Naveen~Nikamanth Appiah~Balaji, Jesse~T. Gonzalez, Negin Heravi, Fahad Islam, Xueting Li, Jean-Baptiste Regli, Kejia Ren, Petr {\v{S}}vec, David~J. Yoon, Matthias M{\"u}ller, and Vladlen Koltun.
\newblock On agentic manipulation policies.
\newblock \emph{arXiv preprint arXiv:2611.09892}, 2026{\natexlab{a}}.

\bibitem[Chen et~al.(2026{\natexlab{b}})Chen, Wang, Chen, Chen, Gao, Tang, Li, Liu, Yao, Li, et~al.]{chen2026lawam}
Jialei Chen, Kai Wang, Kang Chen, Shuaihang Chen, Feng Gao, Wenhao Tang, Zhiyuan Li, Weilin Liu, Zhuyu Yao, Boxun Li, et~al.
\newblock Lawam: Latent world action models for efficient dynamics-aware robot policies.
\newblock \emph{arXiv preprint arXiv:2606.15768}, 2026{\natexlab{b}}.

\bibitem[Chen et~al.(2026{\natexlab{c}})Chen, Hadfield, Zook, Uy, Song, Coumans, Yang, Ladhak, Qu, Birchfield, Tremblay, and Blukis]{chen2026volo}
Siyi Chen, Hugo Hadfield, Alex Zook, Mikaela~Angelina Uy, Chan~Hee Song, Erwin Coumans, Xuning Yang, Faisal Ladhak, Qing Qu, Stan Birchfield, Jonathan Tremblay, and Valts Blukis.
\newblock Volo: A physical orchestrator for open-vocabulary long-horizon manipulation.
\newblock \emph{arXiv preprint arXiv:2606.07723}, 2026{\natexlab{c}}.

\bibitem[Chen et~al.(2026{\natexlab{d}})Chen, Bai, Cao, Zeng, Lin, Lin, Liang, Ma, Huang, and Shou]{chen2026show}
Yanzhe Chen, Zechen Bai, Zhijun Cao, Wenzheng Zeng, Kevin~Qinghong Lin, Yiqi Lin, Guoqiang Liang, Kevin~Yuchen Ma, Qiming Huang, and Mike~Zheng Shou.
\newblock Show-harness: Just a vlm agent can play robots.
\newblock \emph{arXiv preprint arXiv:2609.10522}, 2026{\natexlab{d}}.

\bibitem[Cheng et~al.(2026)Cheng, Yi, Fang, Zhang, Feng, Li, Zhuang, Wang, Yang, Song, et~al.]{cheng2026context}
Dongzhou Cheng, Taoran Yi, Ye Fang, Xingwu Zhang, Fan Feng, Yixuan Li, Gengxiong Zhuang, Rongze Wang, Shuai Yang, Wei Song, et~al.
\newblock In-context robot learning with vlm agents.
\newblock \emph{arXiv preprint arXiv:2609.19138}, 2026.

\bibitem[Driess et~al.(2023)Driess, Xia, Sajjadi, Lynch, Chowdhery, Ichter, Wahid, Tompson, Vuong, Yu, Huang, Chebotar, Sermanet, Duckworth, Levine, Vanhoucke, Hausman, Toussaint, Greff, Zeng, Mordatch, and Florence]{driess2023palme}
Danny Driess, Fei Xia, Mehdi S.~M. Sajjadi, Corey Lynch, Aakanksha Chowdhery, Brian Ichter, Ayzaan Wahid, Jonathan Tompson, Quan Vuong, Tianhe Yu, Wenlong Huang, Yevgen Chebotar, Pierre Sermanet, Daniel Duckworth, Sergey Levine, Vincent Vanhoucke, Karol Hausman, Marc Toussaint, Klaus Greff, Andy Zeng, Igor Mordatch, and Pete Florence.
\newblock Palm-e: An embodied multimodal language model.
\newblock \emph{arXiv preprint arXiv:2303.03378}, 2023.

\bibitem[Driess et~al.(2025)Driess, Springenberg, Ichter, Yu, Li-Bell, Pertsch, Ren, Walke, Vuong, Shi, and Levine]{driess2025insulating}
Danny Driess, Jost~Tobias Springenberg, Brian Ichter, Lili Yu, Adrian Li-Bell, Karl Pertsch, Allen~Z. Ren, Homer Walke, Quan Vuong, Lucy~Xiaoyang Shi, and Sergey Levine.
\newblock Knowledge insulating vision-language-action models: Train fast, run fast, generalize better.
\newblock \emph{arXiv preprint arXiv:2505.23705}, 2025.

\bibitem[Forest~Labs(2026)]{bfl2026flux3action}
Black Forest~Labs.
\newblock Flux 3 action, 2026.

\bibitem[Fu et~al.(2026)Fu, Yu, El-Refai, Kou, Xue, Huang, Xiao, Wang, Li, Shi, Wu, Sastry, Zhu, Goldberg, and Fan]{fu2025capx}
Max Fu, Justin Yu, Karim El-Refai, Ethan Kou, Haoru Xue, Huang Huang, Wenli Xiao, Guanzhi Wang, Fei-Fei Li, Guanya Shi, Jiajun Wu, Shankar Sastry, Yuke Zhu, Ken Goldberg, and Jim Fan.
\newblock {CaP-X}: A framework for benchmarking and improving coding agents for robot manipulation.
\newblock \emph{arXiv preprint arXiv:2603.22435}, 2026.

\bibitem[Hirose et~al.(2026)Hirose, Glossop, Shah, and Levine]{hirose2026asyncvla}
Noriaki Hirose, Catherine Glossop, Dhruv Shah, and Sergey Levine.
\newblock Asyncvla: An asynchronous vla for fast and robust navigation on the edge.
\newblock \emph{arXiv preprint arXiv:2602.13476}, 2026.

\bibitem[Hu et~al.(2026)Hu, Zaech, Nikolov, Yao, Dey, Albanese, Detry, Van~Gool, and Paudel]{hu2026ar}
Yutong Hu, Jan-Nico Zaech, Nikolay Nikolov, Yuanqi Yao, Sombit Dey, Giuliano Albanese, Renaud Detry, Luc Van~Gool, and Danda Paudel.
\newblock Ar-vla: True autoregressive action expert for vision-language-action models.
\newblock \emph{arXiv preprint arXiv:2603.10126}, 2026.

\bibitem[Huang et~al.(2026{\natexlab{a}})Huang, Li, Hu, Qi, Xiao, Coates, Cao, Zhang, and Zhang]{huang2026roboharness}
Jinbang Huang, Zhiyuan Li, Yuanzhao Hu, Ran Qi, Yixin Xiao, Mark Coates, Tongtong Cao, Zhanguang Zhang, and Yingxue Zhang.
\newblock Roboharness: Memory-driven orchestration of heterogeneous robot policies for long-horizon planning.
\newblock \emph{arXiv preprint arXiv:2607.18060}, 2026{\natexlab{a}}.

\bibitem[Huang et~al.(2026{\natexlab{b}})Huang, Zhang, Liu, Song, Tang, and Ma]{huang2026tic}
Zhiyu Huang, Yun Zhang, Johnson Liu, Rui Song, Chen Tang, and Jiaqi Ma.
\newblock {TIC-VLA}: A think-in-control vision-language-action model for robot navigation in dynamic environments.
\newblock \emph{arXiv preprint arXiv:2602.02459}, 2026{\natexlab{b}}.

\bibitem[Jiang et~al.(2025)Jiang, Cheng, Ding, Gao, and Qi]{jiang2026asyncflow}
Yuhua Jiang, Shuang Cheng, Yan Ding, Feifei Gao, and Biqing Qi.
\newblock Asyncvla: Asynchronous flow matching for vision-language-action models.
\newblock \emph{arXiv preprint arXiv:2511.14148}, 2025.

\bibitem[Khazatsky et~al.(2024)Khazatsky, Pertsch, Nair, Balakrishna, Dasari, Karamcheti, Nasiriany, Srirama, Chen, Ellis, et~al.]{khazatsky2024droid}
Alexander Khazatsky, Karl Pertsch, Suraj Nair, Ashwin Balakrishna, Sudeep Dasari, Siddharth Karamcheti, Soroush Nasiriany, Mohan~Kumar Srirama, Lawrence~Yunliang Chen, Kirsty Ellis, et~al.
\newblock Droid: A large-scale in-the-wild robot manipulation dataset.
\newblock In \emph{Robotics: Science and Systems}, 2024.

\bibitem[Kim et~al.(2024)Kim, Pertsch, Karamcheti, Xiao, Balakrishna, Nair, Rafailov, Foster, Lam, Sanketi, et~al.]{kim2024openvla}
Moo~Jin Kim, Karl Pertsch, Siddharth Karamcheti, Ted Xiao, Ashwin Balakrishna, Suraj Nair, Rafael Rafailov, Ethan Foster, Grace Lam, Pannag Sanketi, et~al.
\newblock Openvla: An open-source vision-language-action model.
\newblock \emph{arXiv preprint arXiv:2406.09246}, 2024.

\bibitem[Kim et~al.(2026)Kim, Gao, Lin, Lin, Ge, Lam, Liang, Song, Liu, Finn, et~al.]{kim2026cosmos}
Moo~Jin Kim, Yihuai Gao, Tsung-Yi Lin, Yen-Chen Lin, Yunhao Ge, Grace Lam, Percy Liang, Shuran Song, Ming-Yu Liu, Chelsea Finn, et~al.
\newblock Cosmos policy: Fine-tuning video models for visuomotor control and planning.
\newblock \emph{arXiv preprint arXiv:2601.16163}, 2026.

\bibitem[{LADDER Team}(2026)]{ladderLeaderboard2026}
{LADDER Team}.
\newblock {LADDER}: {RoboLab-Long} leaderboard, november 15 snapshot.
\newblock \url{https://ladderrobotics.ai/\#leaderboard}, 2026.

\bibitem[Li et~al.(2024)Li, Liang, Wang, Luo, Chen, Liao, Wei, Deng, Xu, Zhang, et~al.]{li2024cogact}
Qixiu Li, Yaobo Liang, Zeyu Wang, Lin Luo, Xi Chen, Mozheng Liao, Fangyun Wei, Yu Deng, Sicheng Xu, Yizhong Zhang, et~al.
\newblock Cogact: A foundational vision-language-action model for synergizing cognition and action in robotic manipulation.
\newblock \emph{arXiv preprint arXiv:2411.19650}, 2024.

\bibitem[Li et~al.(2026)Li, Gong, Thompson, Javanmardi, Zhang, Zeng, Liu, Wang, Guo, and Tsukada]{li2026think}
Yun Li, Jiachen Gong, Simon Thompson, Ehsan Javanmardi, Qunli Zhang, Zifan Zeng, Shiming Liu, Peng Wang, Zixuan Guo, and Manabu Tsukada.
\newblock Think at 5 hz, act at 20 hz: Asynchronous fast-slow vision-language-action inference for closed-loop driving.
\newblock \emph{arXiv preprint arXiv:2607.15621}, 2026.

\bibitem[Liang et~al.(2022)Liang, Huang, Xia, Xu, Hausman, Ichter, Florence, and Zeng]{codeaspolicies2022}
Jacky Liang, Wenlong Huang, Fei Xia, Peng Xu, Karol Hausman, Brian Ichter, Pete Florence, and Andy Zeng.
\newblock Code as policies: Language model programs for embodied control.
\newblock \emph{arXiv preprint arXiv:2209.07753}, 2022.

\bibitem[Liao et~al.(2025)Liao, Kim, Smith, Agha-mohammadi, and Omidshafiei]{liao2025delay}
Aileen Liao, Dong-Ki Kim, Max~Olan Smith, Ali-akbar Agha-mohammadi, and Shayegan Omidshafiei.
\newblock Delay-aware diffusion policy: Bridging the observation-execution gap in dynamic tasks.
\newblock \emph{arXiv preprint arXiv:2512.07697}, 2025.

\bibitem[Lipman et~al.(2024)Lipman, Havasi, Holderrieth, Shaul, Le, Karrer, Chen, Lopez-Paz, Ben-Hamu, and Gat]{lipman2024flow}
Yaron Lipman, Marton Havasi, Peter Holderrieth, Neta Shaul, Matt Le, Brian Karrer, Ricky~TQ Chen, David Lopez-Paz, Heli Ben-Hamu, and Itai Gat.
\newblock Flow matching guide and code.
\newblock \emph{arXiv preprint arXiv:2412.06264}, 2024.

\bibitem[{NVIDIA}(2025)]{nvidia2025gr00tn16}
{NVIDIA}.
\newblock {GR00T N1.6-3B}: Model card.
\newblock \url{https://huggingface.co/nvidia/GR00T-N1.6-3B}, 2025.

\bibitem[{NVIDIA}(2026{\natexlab{a}})]{cosmos3}
{NVIDIA}.
\newblock {Cosmos 3}: Omnimodal world models for physical {AI}.
\newblock \emph{arXiv preprint arXiv:2606.02800}, 2026{\natexlab{a}}.

\bibitem[{NVIDIA}(2026{\natexlab{b}})]{robolabLeaderboard2026}
{NVIDIA}.
\newblock {RoboLab-120} leaderboard, november 15 snapshot.
\newblock \url{https://research.nvidia.com/labs/srl/projects/robolab/leaderboard.html}, 2026{\natexlab{b}}.

\bibitem[{NVIDIA} et~al.(2025){NVIDIA}, Bjorck, Castañeda, Cherniadev, Da, Ding, Fan, Fang, Fox, Hu, Huang, Jang, Jiang, Kautz, Kundalia, Lao, Li, Lin, Lin, Liu, Llontop, Magne, Mandlekar, Narayan, Nasiriany, Reed, Tan, Wang, Wang, Wang, Wang, Xiang, Xie, Xu, Xu, Ye, Yu, Zhang, Zhang, Zhao, Zheng, and Zhu]{nvidia2025gr00tn1openfoundation}
{NVIDIA}, Johan Bjorck, Fernando Castañeda, Nikita Cherniadev, Xingye Da, Runyu Ding, Linxi~"Jim" Fan, Yu Fang, Dieter Fox, Fengyuan Hu, Spencer Huang, Joel Jang, Zhenyu Jiang, Jan Kautz, Kaushil Kundalia, Lawrence Lao, Zhiqi Li, Zongyu Lin, Kevin Lin, Guilin Liu, Edith Llontop, Loic Magne, Ajay Mandlekar, Avnish Narayan, Soroush Nasiriany, Scott Reed, You~Liang Tan, Guanzhi Wang, Zu Wang, Jing Wang, Qi Wang, Jiannan Xiang, Yuqi Xie, Yinzhen Xu, Zhenjia Xu, Seonghyeon Ye, Zhiding Yu, Ao Zhang, Hao Zhang, Yizhou Zhao, Ruijie Zheng, and Yuke Zhu.
\newblock Gr00t n1: An open foundation model for generalist humanoid robots, 2025.

\bibitem[{Octo Model Team} et~al.(2024){Octo Model Team}, Ghosh, Walke, Pertsch, Black, Mees, Dasari, Hejna, Xu, Luo, Kreiman, Tan, Chen, Sanketi, Vuong, Xiao, Sadigh, Finn, and Levine]{octo_2023}
{Octo Model Team}, Dibya Ghosh, Homer Walke, Karl Pertsch, Kevin Black, Oier Mees, Sudeep Dasari, Joey Hejna, Charles Xu, Jianlan Luo, Tobias Kreiman, {You Liang} Tan, Lawrence~Yunliang Chen, Pannag Sanketi, Quan Vuong, Ted Xiao, Dorsa Sadigh, Chelsea Finn, and Sergey Levine.
\newblock Octo: An open-source generalist robot policy.
\newblock In \emph{Proceedings of Robotics: Science and Systems}, Delft, Netherlands, 2024.

\bibitem[Padalkar et~al.(2023)Padalkar, Pooley, Jain, Bewley, Herzog, Irpan, Khazatsky, Rai, Singh, Brohan, et~al.]{open_x_embodiment_rt_x_2023}
Abhishek Padalkar, Acorn Pooley, Ajinkya Jain, Alex Bewley, Alex Herzog, Alex Irpan, Alexander Khazatsky, Anant Rai, Anikait Singh, Anthony Brohan, et~al.
\newblock Open x-embodiment: Robotic learning datasets and rt-x models.
\newblock \emph{arXiv preprint arXiv:2310.08864}, 2023.

\bibitem[Pertsch et~al.(2025)Pertsch, Stachowicz, Ichter, Driess, Nair, Vuong, Mees, Finn, and Levine]{pertsch2025fast}
Karl Pertsch, Kyle Stachowicz, Brian Ichter, Danny Driess, Suraj Nair, Quan Vuong, Oier Mees, Chelsea Finn, and Sergey Levine.
\newblock Fast: Efficient action tokenization for vision-language-action models.
\newblock \emph{arXiv preprint arXiv:2501.09747}, 2025.

\bibitem[Petkov et~al.(2026)Petkov, Albanese, Nikolov, Van~Gool, Tremblay, and Paudel]{ladder2026}
Petko Petkov, Giuliano Albanese, Nikolay Nikolov, Luc Van~Gool, Jonathan Tremblay, and Danda~Pani Paudel.
\newblock {LADDER}: A long-horizon robot manipulation benchmark.
\newblock \emph{arXiv preprint arXiv:2610.18053}, 2026.

\bibitem[Qu et~al.(2025)Qu, Song, Chen, Yao, Ye, Ding, Wang, Gu, Zhao, Wang, et~al.]{qu2025spatialvla}
Delin Qu, Haoming Song, Qizhi Chen, Yuanqi Yao, Xinyi Ye, Yan Ding, Zhigang Wang, JiaYuan Gu, Bin Zhao, Dong Wang, et~al.
\newblock Spatialvla: Exploring spatial representations for visual-language-action model.
\newblock \emph{arXiv preprint arXiv:2501.15830}, 2025.

\bibitem[Reuss et~al.(2025)Reuss, Zhou, R{\"u}hle, Ya{\u{g}}murlu, Otto, and Lioutikov]{reuss2025flower}
Moritz Reuss, Hongyi Zhou, Marcel R{\"u}hle, {\"O}mer~Erdin{\c{c}} Ya{\u{g}}murlu, Fabian Otto, and Rudolf Lioutikov.
\newblock {FLOWER}: Democratizing generalist robot policies with efficient vision-language-action flow policies.
\newblock In \emph{7th Robot Learning Workshop: Towards Robots with Human-Level Abilities}, 2025.

\bibitem[Shen et~al.(2026)Shen, Kumar, Chintalapudi, Lindeborg, Wang, Watson, Hu, Cao, Jayaraman, Kaelbling, and Lozano-P{\'e}rez]{shen2026tiptop}
William Shen, Nishanth Kumar, Sahit Chintalapudi, Ryan Lindeborg, Jie Wang, Christopher Watson, Edward~S. Hu, Jing Cao, Dinesh Jayaraman, Leslie~Pack Kaelbling, and Tom{\'a}s Lozano-P{\'e}rez.
\newblock {TiPToP}: A modular open-vocabulary robot manipulation system that plans.
\newblock \emph{arXiv preprint arXiv:2603.09971}, 2026.

\bibitem[Shi et~al.(2025)Shi, Yang, Chao, Wan, Shao, Lei, Qian, Le, Chaudhari, Daniilidis, et~al.]{shi2025maestro}
Junyao Shi, Rujia Yang, Kaitian Chao, Selina~Bingqing Wan, Yifei Shao, Jiahui Lei, Jianing Qian, Long Le, Pratik Chaudhari, Kostas Daniilidis, et~al.
\newblock Maestro: Orchestrating robotics modules with vision-language models for zero-shot generalist robots.
\newblock In \emph{NeurIPS 2025 Workshop on Space in Vision, Language, and Embodied AI}, 2025.

\bibitem[Sim{\'e}oni et~al.(2025)]{simeoni2025dinov3}
Oriane Sim{\'e}oni et~al.
\newblock {DINOv3}.
\newblock \emph{arXiv preprint arXiv:2508.10104}, 2025.

\bibitem[Tang et~al.(2025)Tang, Sun, Zhao, Yang, Lin, Zhang, Hou, Lu, Liu, and Han]{tang2025vlash}
Jiaming Tang, Yufei Sun, Yilong Zhao, Shang Yang, Yujun Lin, Zhuoyang Zhang, James Hou, Yao Lu, Zhijian Liu, and Song Han.
\newblock Vlash: Real-time vlas via future-state-aware asynchronous inference.
\newblock \emph{arXiv preprint arXiv:2512.01031}, 2025.

\bibitem[Vanjani et~al.(2026)Vanjani, Li, Suliga, Reuss, Geraci, Jiang, and Lioutikov]{vanjani2026dam}
Pankhuri Vanjani, Zhuoyue Li, Jakub Suliga, Moritz Reuss, Gianluca Geraci, Xinkai Jiang, and Rudolf Lioutikov.
\newblock Dam-vla: Decoupled asynchronous multimodal vision language action model.
\newblock \emph{arXiv preprint arXiv:2606.12105}, 2026.

\bibitem[Walke et~al.(2023)Walke, Black, Lee, Kim, Du, Zheng, Zhao, Hansen-Estruch, Vuong, He, Myers, Fang, Finn, and Levine]{walke2023bridgedata}
Homer Walke, Kevin Black, Abraham Lee, Moo~Jin Kim, Max Du, Chongyi Zheng, Tony Zhao, Philippe Hansen-Estruch, Quan Vuong, Andre He, Vivek Myers, Kuan Fang, Chelsea Finn, and Sergey Levine.
\newblock Bridgedata v2: A dataset for robot learning at scale.
\newblock In \emph{Proceedings of the Conference on Robot Learning (CoRL)}, 2023.

\bibitem[Xie et~al.(2026)Xie, Wen, Zheng, Chen, Hong, Diao, and Liu]{xie2026dynamicvla}
Haozhe Xie, Beichen Wen, Jiarui Zheng, Zhaoxi Chen, Fangzhou Hong, Haiwen Diao, and Ziwei Liu.
\newblock {DynamicVLA:} a vision-language-action model for dynamic object manipulation.
\newblock In \emph{NeurIPS}, 2026.

\bibitem[Yan et~al.(2026)Yan, Wang, Liu, Meng, Yang, Yao, Feng, Ma, Zhao, and Han]{yan2026acting}
Shenhao Yan, Ge Wang, Qi Liu, Weilin Meng, Jiahao Yang, Chengsi Yao, Fan Feng, Xiaoguang Ma, Yiming Zhao, and Yatong Han.
\newblock Acting while understanding: Asynchronous semantic-action decoupling for real-time vision-language-action models, 2026.
\newblock arXiv:2606.15285.

\bibitem[Yang et~al.(2026{\natexlab{a}})Yang, Khetan, and Zhang]{yang2026astranomy}
Jenai~Xuning Yang, Ashna Khetan, and Shun Zhang.
\newblock Astranomy: The study of {GPT-6 Astra} as a native robot policy.
\newblock \url{https://research.nvidia.com/labs/srl/projects/astranomy/}, 2026{\natexlab{a}}.
\newblock Blog post.

\bibitem[Yang et~al.(2026{\natexlab{b}})Yang, Dagli, Zook, Hadfield, Goyal, Birchfield, Ramos, and Tremblay]{yang2026robolab}
Xuning Yang, Rishit Dagli, Alex Zook, Hugo Hadfield, Ankit Goyal, Stan Birchfield, Fabio Ramos, and Jonathan Tremblay.
\newblock {RoboLab: A High-Fidelity Simulation Benchmark for Analysis of Task Generalist Policies}.
\newblock In \emph{Proceedings of Robotics: Science and Systems}, Sydney, Australia, 2026{\natexlab{b}}.

\bibitem[Ye et~al.(2026)Ye, Ge, Zheng, Gao, Yu, Kurian, Indupuru, Tan, Zhu, Xiang, Malik, Lee, Liang, Ranawaka, Gu, Xu, Wang, Hu, Narayan, Bjorck, Wang, Kim, Niu, Zheng, Xie, Wu, Wang, Julian, Xu, Du, Chebotar, Reed, Kautz, Zhu, Fan, and Jang]{ye2026worldactionmodelszeroshot}
Seonghyeon Ye, Yunhao Ge, Kaiyuan Zheng, Shenyuan Gao, Sihyun Yu, George Kurian, Suneel Indupuru, You~Liang Tan, Chuning Zhu, Jiannan Xiang, Ayaan Malik, Kyungmin Lee, William Liang, Nadun Ranawaka, Jiasheng Gu, Yinzhen Xu, Guanzhi Wang, Fengyuan Hu, Avnish Narayan, Johan Bjorck, Jing Wang, Gwanghyun Kim, Dantong Niu, Ruijie Zheng, Yuqi Xie, Jimmy Wu, Qi Wang, Ryan Julian, Danfei Xu, Yilun Du, Yevgen Chebotar, Scott Reed, Jan Kautz, Yuke Zhu, Linxi~"Jim" Fan, and Joel Jang.
\newblock World action models are zero-shot policies, 2026.

\bibitem[Zawalski et~al.(2024)Zawalski, Chen, Pertsch, Mees, Finn, and Levine]{zawalski2024ecot}
Micha{\l} Zawalski, William Chen, Karl Pertsch, Oier Mees, Chelsea Finn, and Sergey Levine.
\newblock Robotic control via embodied chain-of-thought reasoning.
\newblock In \emph{8th Annual Conference on Robot Learning}, 2024.

\bibitem[Zha et~al.(2026)Zha, Hancock, Zhang, Yin, Huang, Shah, Ren, and Majumdar]{zha2026lap}
Lihan Zha, Asher~J Hancock, Mingtong Zhang, Tenny Yin, Yixuan Huang, Dhruv Shah, Allen~Z Ren, and Anirudha Majumdar.
\newblock Lap: Language-action pre-training enables zero-shot cross-embodiment transfer.
\newblock \emph{arXiv preprint arXiv:2602.10556}, 2026.

\bibitem[Zhao et~al.(2025)Zhao, Lu, Kim, Fu, Zhang, Wu, Li, Ma, Han, Finn, et~al.]{zhao2025cotvla}
Qingqing Zhao, Yao Lu, Moo~Jin Kim, Zipeng Fu, Zhuoyang Zhang, Yecheng Wu, Zhaoshuo Li, Qianli Ma, Song Han, Chelsea Finn, et~al.
\newblock Cot-vla: Visual chain-of-thought reasoning for vision-language-action models.
\newblock In \emph{Proceedings of the Computer Vision and Pattern Recognition Conference}, pages 1702--1713, 2025.

\bibitem[Zhou et~al.(2024)Zhou, Pan, LeCun, and Pinto]{zhou2024dino}
Gaoyue Zhou, Hengkai Pan, Yann LeCun, and Lerrel Pinto.
\newblock Dino-wm: World models on pre-trained visual features enable zero-shot planning.
\newblock \emph{arXiv preprint arXiv:2411.04983}, 2024.

\bibitem[Zitkovich et~al.(2023)Zitkovich, Yu, Xu, Xu, Xiao, Xia, Wu, Wohlhart, Welker, Wahid, Vuong, Vanhoucke, Tran, Soricut, Singh, Singh, Sermanet, Sanketi, Salazar, Ryoo, Reymann, Rao, Pertsch, Mordatch, Michalewski, Lu, Levine, Lee, Lee, Leal, Kuang, Kalashnikov, Julian, Joshi, Irpan, Ichter, Hsu, Herzog, Hausman, Gopalakrishnan, Fu, Florence, Finn, Dubey, Driess, Ding, Choromanski, Chen, Chebotar, Carbajal, Brown, Brohan, Arenas, and Han]{zitkovich2023rt2}
Brianna Zitkovich, Tianhe Yu, Sichun Xu, Peng Xu, Ted Xiao, Fei Xia, Jialin Wu, Paul Wohlhart, Stefan Welker, Ayzaan Wahid, Quan Vuong, Vincent Vanhoucke, Huong Tran, Radu Soricut, Anikait Singh, Jaspiar Singh, Pierre Sermanet, Pannag~R. Sanketi, Grecia Salazar, Michael~S. Ryoo, Krista Reymann, Kanishka Rao, Karl Pertsch, Igor Mordatch, Henryk Michalewski, Yao Lu, Sergey Levine, Lisa Lee, Tsang-Wei~Edward Lee, Isabel Leal, Yuheng Kuang, Dmitry Kalashnikov, Ryan Julian, Nikhil~J. Joshi, Alex Irpan, Brian Ichter, Jasmine Hsu, Alexander Herzog, Karol Hausman, Keerthana Gopalakrishnan, Chuyuan Fu, Pete Florence, Chelsea Finn, Kumar~Avinava Dubey, Danny Driess, Tianli Ding, Krzysztof~Marcin Choromanski, Xi Chen, Yevgen Chebotar, Justice Carbajal, Noah Brown, Anthony Brohan, Montserrat~Gonzalez Arenas, and Kehang Han.
\newblock Rt-2: Vision-language-action models transfer web knowledge to robotic control.
\newblock In \emph{Proceedings of The 7th Conference on Robot Learning}, pages 2165--2183. PMLR, 2023.

\bibitem[Zou et~al.(2025)Zou, Zeng, Nong, Li, Liu, Yang, Ling, Li, and Ma]{zou2025duocore}
Teqiang Zou, Hongliang Zeng, Yuxuan Nong, Yifan Li, Kehui Liu, Haotian Yang, Xinyang Ling, Xin Li, and Lianyang Ma.
\newblock Asynchronous fast-slow vision-language-action policies for whole-body robotic manipulation.
\newblock \emph{arXiv preprint arXiv:2512.20188}, 2025.

\end{thebibliography}
\end{document}